%% file: main.tex
\def\mytitle{Second-Order Stochastic Optimization for Machine Learning in Linear Time}

\def\myauthors{Naman Agarwal\footnote{namana@cs.princeton.edu,
Computer Science, Princeton University}\hspace{1.5cm} Brian
Bullins\footnote{bbullins@cs.princeton.edu, Computer Science, Princeton
University}\hspace{1.5cm} Elad Hazan\footnote{ehazan@cs.princeton.edu,
Computer Science, Princeton University}}

\documentclass[twoside,11pt]{article}
\usepackage[fullpage,nousetoc,hylinks]{paper}

\theoremstyle{plain}
\usepackage{amsmath}
\usepackage{algorithm}
\usepackage{algorithmic}

\usepackage{makecell}

\include{defs}

\BEGINDOC
\begin{titlepage}
\makeheader

\begin{abstract}
First-order stochastic methods are the state-of-the-art in large-scale machine learning optimization owing to efficient per-iteration complexity. Second-order methods, while able to provide faster convergence, have been much less explored due to the high cost of computing the second-order information. In this paper we develop second-order stochastic methods for optimization problems in machine learning that match the per-iteration cost of gradient based methods, and in certain settings improve upon the overall running time over popular first-order methods. Furthermore, our algorithm has the desirable property of being implementable in time linear in the sparsity of the input data. 
\end{abstract}
\end{titlepage}

\section{Introduction}

In recent literature stochastic first-order optimization has taken the stage as
the primary workhorse for training learning models, due in large part to its affordable computational costs which are linear (in the data representation) per 
iteration. The main research effort devoted to improving the convergence rates of first-order methods have introduced elegant ideas and algorithms in recent years, including adaptive 
regularization \cite{adagrad}, variance reduction \cite{SVRG,SAGA}, dual
coordinate ascent \cite{SDCA}, and many more.

In contrast, second-order methods have typically been much less explored in large scale machine learning (ML) applications due to their prohibitive computational cost per iteration 
which requires computation of the Hessian in addition to a matrix
inversion. These operations are infeasible for
large scale problems in high dimensions. 

In this paper we propose a family of novel second-order algorithms, LiSSA (Linear time
Stochastic Second-Order Algorithm) for convex
optimization that attain fast convergence rates while also allowing for an implementation with linear time per-iteration cost, matching the running time of the best known gradient-based methods. Moreover, in the setting where the number of training examples $m$ is much larger than the underlying dimension $d$, we show that our algorithm has provably faster running time than the best known gradient-based methods.

Formally, the main optimization problem we are concerned with is the empirical risk minimization (ERM) problem: 
$$ \min_{\x \in \reals^d} f(\x)  = \min_{\x \in \reals^d}
\left\{ \frac{1}{m} \sum_{k=1}^{m} f_k(\x)  + R(\x) \right \} $$
where each $f_k(\x)$ is a convex function and $R(\x)$ is a
convex regularizer. The above optimization problem is the standard objective minimized in most supervised learning settings. Examples include logistic regression, SVMs, etc. A common aspect of many applications of ERM in machine learning is that the loss function $f_i(\x)$ is of the form $l(\x^\top \vv_i, y_i)$ where $(\vv_i, y_i)$ is the $i^{th}$ training example-label pair. We call such functions generalized linear models (GLM) and will restrict our attention to this case. We will assume that the regularizer is an $\ell_2$ regularizer,\footnote{It can be seen that some of these assumptions can be relaxed in our analyses, but since these choices are standard in ML,  we make these assumptions to simplify the discourse.} typically $\|\x\|^2$.

Our focus is second-order optimization methods (Newton's method), where in each iteration, the underlying principle is to move to the minimizer of the second-order Taylor approximation at any point. Throughout the paper, we will let $\hessinv f(\x) \defeq \left[\hess f(\x)\right]^{-1}$. The update of Newton's method at a point $\x_t$ is then given by
\begin{equation}
\label{eqn:newtonupdate}
\x_{t+1} = \x_{t} - \hessinv f(\x_{t}) \nabla f(\x_{t}).
\end{equation}
Certain desirable properties of Newton's method include the fact that its updates are independent of the choice of coordinate system and that the Hessian provides the necessary regularization based on the curvature at the present point. Indeed, Newton's method can be shown to eventually achieve quadratic convergence  \cite{NesterovBook}. Although Newton's method comes with good theoretical guarantees, the
complexity per step grows roughly as $\Omega(md^2 + d^\omega)$ (the former term for computing
the Hessian and the latter for inversion, where $\omega \approx 2.37$ is the matrix multiplication constant), making it prohibitive in practice. Our main contribution is a suite of algorithms, each of which performs an approximate Newton update based on stochastic Hessian information and is implementable in linear $O(d)$ time.  These algorithms match and improve over the performance of first-order methods in theory and give promising results as an optimization method on real world data sets. In the following we give  a summary of our results. We propose two algorithms, LiSSA and LiSSA-Sample.

\textit{LiSSA}: Algorithm \ref{alg:stochasticnewton} is a practical stochastic second-order algorithm based on a novel estimator of the Hessian inverse, leading to an efficient approximate Newton step (Equation \ref{eqn:newtonupdate}). The estimator is based on the well known Taylor approximation of the inverse (Fact \ref{eqn:taylorexpansion}) and is described formally in Section \ref{sec:estimators}. We prove the following informal theorem about LiSSA.

\begin{theorem}[Informal]
      LiSSA returns a point $\x_t$ such that $f(\x_t) \leq \min_{\x^*} f(\x^*) +
      \varepsilon$ in total time \[\tilde{O}\left(\left(m +
      S_1\kappa \right)d \log \left(\frac{1}{
      \varepsilon}\right) \right) \]
      where $\kappa$ is the underlying condition number of the problem and $S_1$ is a bound on the variance of the estimator.
\end{theorem}

  The precise version of the above theorem appears as Theorem \ref{thm:mainthm}. In theory, the best bound we can show for $S_1$ is $O(\kappa^2)$; however, in our experiments we observe that setting $S_1$ to be a small constant (often 1) is sufficient. We conjecture that $S_1$ can be improved to $O(1)$ and leave this for future work. If indeed $S_1$ can be improved to $O(1)$ (as is indicated by our experiments), LiSSA enjoys a convergence rate comparable to first-order methods.  We provide a detailed comparison of our results with existing first-order and second-order methods in Section \ref{sec:relatedwork}. Moreover, in Section \ref{sec:experiments} we present experiments on real world data sets that demonstrate that LiSSA as an optimization method performs well as compared to popular first-order methods. We also show that LiSSA runs in time proportional to input sparsity, making it an attractive method for high-dimensional sparse data. 

\textit{LiSSA-Sample}: This variant brings together efficient first-order algorithms with matrix sampling techniques \cite{PengRowSampling,YinTatPaper} to achieve better runtime guarantees than the state-of-the-art in convex optimization for machine learning in the regime when $m > d$. Specifically, we prove the following theorem:
      \begin{theorem}[Informal]
      LiSSA-Sample returns a point $\x_t$ such that $f(\x_t) \leq \min_{\x^*} f(\x^*) +
      \varepsilon$ in total time \[\tilde{O}\left(m +
      \sqrt{\kappa d}\right)d \log^2 \left(\frac{1}{
      \varepsilon}\right)\log \log \left(\frac{1}{
      \varepsilon}\right). \]
      \end{theorem}
The above result improves upon the best known running time for first-order methods achieved by acceleration when we are in the setting where $\kappa > m >> d$. We discuss the implication of our bounds and further work in Section \ref{sec:discussion}.

In all of our results stated above $\kappa$ corresponds to the condition number of the underlying problem. In particular we assume some strong convexity for the underlying problem. This is a standard assumption which is usually enforced by the addition of the $\ell_2$ regularizer. In stating our results formally we stress on the  nuances between different notions of the condition number (ref. Section \ref{sec:conditionnumberappendix}), and we state our results precisely with respect to these notions. In general, all of our generalization/relaxations of the condition number are smaller than $\frac{1}{\lambda}$ where $\lambda$ is the regularization parameter, and this is usually taken to be the condition number of the problem. The condition of strong convexity has been relaxed in literature by introducing proximal methods. It is an interesting direction to adapt our results in those setting which we leave for future work.

We also remark that all of our results focus on the very high accuracy regime. In general the benefits of linear convergence and second-order methods can be seen to be effective only when considerably small error is required. This is also the case for recent advances in fast first-order methods where their improvement over stochastic gradient descent becomes apparent only in the high accuracy regime. Our experiments also demonstrate that second-order methods can improve upon fast first-order methods in the regime of very high accuracy.  While it is possible that this regime is less interesting for generalization, in this paper we focus on the optimization problem itself. 

We further consider the special case when
the function $f$ is self-concordant. Self-concordant functions are a
sub-class of convex functions which have been extensively studied in convex
optimization literature in the context of interior point methods
\cite{NemirovskiBook}. For self-concordant functions we propose an algorithm (Algorithm \ref{alg:ellipsoidcover}) which achieves linear convergence with running time guarantees independent of the condition number. We prove the formal running time guarantee as Theorem \ref{thm:selfconctheorem}.

We believe our main contribution to be a demonstration of the fact that second-order methods are comparable to, or even better than, first-order methods in the large data regime, in both theory and practice.

\subsection{Overview of Techniques}

\textit{LiSSA}: The key idea underlying LiSSA is the use of the Taylor expansion to construct a natural estimator of the Hessian inverse. Indeed, as can be seen from the description of the estimator in Section \ref{sec:estimators}, the estimator we construct becomes unbiased in the limit as we include additional terms in the series. We note that this is not the case with estimators that were considered in previous works such as that of \cite{newsamp}, and so we therefore consider our estimator to be more natural. In the implementation of the algorithm we achieve the optimal bias/variance trade-off by truncating the series appropriately.

An important observation underlying our linear time $O(d)$ step is that for GLM functions, $\hess f_i(\x)$ has the form $\alpha \bv_i\bv_i^\top$ where $\alpha$ is a scalar dependent on $\bv_i^\top\x$. A single step of LiSSA requires us to efficiently compute $\hess f_i(\x) \bb$ for a given vector $\bb$. In this case it can be seen that the matrix-vector product reduces to a vector-vector product, giving us an $O(d)$ time update. 

\textit{LiSSA-Sample}: LiSSA-Sample is based on Algorithm \ref{alg:SVRGLiSSA}, which represents a general family of algorithms that couples the quadratic minimization view of Newton's method with any efficient first-order method. In essence, Newton's method allows us to reduce (up to $\log \log$ factors) the optimization of a general convex function to solving intermediate quadratic or ridge regression problems. Such a reduction is useful in two ways.

First, as we demonstrate through our algorithm LiSSA-Sample, the quadratic nature of ridge regression problems allows us to leverage powerful sampling techniques, leading to an improvement over the running time of the best known accelerated first-order method. On a high level this improvement comes from the fact that when solving a system of $m$ linear equations in $d$ dimensions, a constant number of passes through the data is enough to reduce the system to $O(d \log(d))$ equations. We carefully couple this principle and the computation required with accelerated first-order methods to achieve the running times for LiSSA-Sample. The result for the quadratic sub-problem (ridge regression) is stated in Theorem \ref{thm:main_quadratic_thm}, and the result for convex optimization is stated in Theorem \ref{thm:LiSSAquadms}.

The second advantage of the reduction to quadratic sub-problems comes from the observation that the intermediate quadratic sub-problems can potentially be better conditioned than the function itself, allowing us a better choice of the step size in practice. We define these local notions of condition number formally in Section \ref{sec:conditionnumberappendix} and summarize the typical benefits for such algorithms in Theorem \ref{thm:generalLiSSA}. In theory this is not a significant improvement; however, in practice we believe that this could be significant and lead to runtime improvements.\footnote{While this is a difficult property to verify experimentally, we conjecture that this is a possible explanation for why LiSSA performs better than first-order methods on certain data sets and ranges of parameters.} 

To achieve the bound for LiSSA-Sample we extend the definition and procedure for sampling via leverage scores described by \cite{YinTatPaper} to the case when the matrix is given as a sum of PSD matrices and not just rank one matrices. We reformulate and reprove the theorems proved by \cite{YinTatPaper} in this context, which may be of independent interest. 

\subsection{Comparison with Related Work}
\label{sec:relatedwork}
In this section we aim to provide a short summary of the key ideas and results underlying optimization methods for large scale machine learning. We divide the summary into three high level principles: first-order gradient-based methods, second-order Hessian-based methods, and quasi-Newton methods. For the sake of brevity we will restrict our summary to results in the case when the objective is strongly convex, which as justified above is usually ensured by the addition of an appropriate regularizer. In such settings the main focus is often to obtain algorithms which have provably linear convergence and fast implementations.   

\textit{First-Order Methods}: First-order methods have dominated the space of optimization algorithms for machine learning owing largely to the fact that they can be implemented in time proportional to the underlying dimension (or sparsity). Gradient descent is known to converge linearly to the optimum with a rate of convergence that is dependent upon the condition number of the objective. In the large data regime, stochastic first-order methods, introduced and analyzed first by \cite{RM51}, have proven especially successful. Stochastic gradient descent (SGD), however, converges sub-linearly even in the strongly convex setting. A significant advancement in terms of the running time of first-order methods was achieved recently by a clever merging of stochastic gradient descent with its full version to provide \textit{variance reduction}. The representative algorithms in this space are SAGA \cite{Bachpaper,SAGA} and SVRG \cite{SVRG, mahdavi}. The key technical achievement of the above algorithms is to relax the running time dependence on $m$ (the number of training examples) and $\kappa$ (the condition number) from a product to a sum. Another algorithm which achieves similar running time guarantees is based on dual coordinate ascent, known as SDCA \cite{SDCA}. 

Further improvements over SAGA, SVRG and SDCA have been obtained by applying the classical idea of \textit{acceleration} emerging from the seminal work of \cite{nesterovacceleration}. The progression of work here includes an accelerated version of SDCA \cite{AccSDCA}; APCG \cite{lin2014accelerated}; Catalyst \cite{Catalyst2015}, which provides a generic framework to accelerate first order algorithms; and Katyusha \cite{Katyusha2016}, which introduces the concept of negative momentum to extend acceleration for variance reduced algorithms beyond the strongly convex setting. The key technical achievement of accelerated methods in general is to reduce the dependence on condition number from linear to a square root. We summarize these results in Table \ref{table:allruntimes}. 

LiSSA places itself naturally into the space of fast first-order methods by having a running time dependence that is comparable to SAGA/SVRG (ref. Table \ref{table:allruntimes}). In LiSSA-Sample we leverage the quadratic structure of the sub-problem for which efficient sampling techniques have been developed in the literature and use accelerated first-order methods to improve the running time in the case when the underlying dimension is much smaller than the number of training examples. Indeed, to the best of our knowledge LiSSA-Sample is the theoretically fastest known algorithm under the condition $m >> d$. Such an improvement seems out of hand for the present first-order methods as it seems to strongly leverage the quadratic nature of the sub-problem to reduce its size. We summarize these results in Table \ref{table:allruntimes}.   

\begin{table}[h]

\begin{center}
\begin{small}
{
\begin{tabular}{|c|c|c|}
\hline

Algorithm & Runtime & Reference \\
\hline
SVRG, SAGA, SDCA  & 
$\left(md + O(\hat{\kappa} d)\right) \log\left(\frac{1}{\epsilon}\right)$ & \makecell{\cite{SVRG}\\\cite{mahdavi}\\\cite{Bachpaper}\\\cite{SDCA}}
\\
\hline
LiSSA & $\left(md + O(\hat{\kappa}_l)S_1 d\right)\log\left(\frac{1}{\epsilon}\right)$ & Corollary \ref{cor:main_corollary} \\
\hline
\makecell{AccSDCA, Catalyst \\Katyusha} & $\tilde{O}\left( md + d\sqrt{\hat{\kappa} m} \right) \log\left(\frac{1}{\epsilon}\right)$ & \makecell{\cite{AccSDCA}\\\cite{Catalyst2015}\\\cite{Katyusha2016}} \\
\hline
LiSSA-Sample & $\tilde{O}\left(md + d\sqrt{\kappa_{sample} d} \right)\log^2\left(\frac{1}{\epsilon}\right)$ & Theorem \ref{thm:LiSSAquadms}\\
\hline
\end{tabular}
}
\end{small}
\end{center}
\caption{Run time comparisons. Refer to Section \ref{sec:conditionnumberappendix} for definitions of the various notions of condition number.}
\label{table:allruntimes}
\end{table}

\textit{Second-Order Methods}: Second-order methods such as Newton's method have classically been used in optimization in many different settings including
development of interior point methods \cite{NemirovskiBook} for general convex programming. The key advantage of Newton's method is that it achieves a linear-quadratic convergence rate. However, naive implementations of Newton's method have two significant issues, namely that the standard analysis requires the full Hessian calculation which costs $O(md^2)$, an expense not suitable
for machine learning applications, and the matrix inversion typically requires $O(d^3)$ time. These issues were addressed recently by the algorithm NewSamp \cite{newsamp} which tackles the first issue by subsampling and the second issue by low-rank projections. We improve upon the work of \cite{newsamp} by defining a more \textit{natural} estimator for the Hessian inverse and by demonstrating that the estimator can be computed in time proportional to $O(d)$. We also point the reader to the works of \cite{MAR10, BCNN11} which incorporate the idea of taking samples of the Hessian; however, these works do not provide precise running time guarantees on their proposed algorithm based on problem specific parameters. Second-order methods have also enjoyed success in the distributed setting \cite{dane}.

\textit{Quasi-Newton Methods}: The expensive computation of the Newton step has also been tackled via estimation of the curvature from the change in gradients. These algorithms are generally known as quasi-Newton methods stemming from the seminal BFGS algorithm \cite{bro70, fle70, gol70, sha70}. The book of \cite{nocedalbook} is an excellent reference for the algorithm and its limited memory variant (L-BFGS). The more recent work in this area has focused on stochastic quasi-Newton methods which were proposed and analyzed in various settings by \cite{SYGSQN,MRSQN,BHNS14}. These works typically achieve sub-linear convergence to the optimum. A significant advancement in this line of work was provided by \cite{SLBFGS} who propose an algorithm based on L-BFGS by incorporating ideas from variance reduction to achieve linear convergence to the optimum in the strongly convex setting. Although the algorithm achieves linear convergence, the running time of the algorithm depends poorly on the condition number (as acknowledged by the authors). Indeed, in applications that interest us, the condition number is not necessarily a constant as is typically assumed to be the case for the theoretical results in \cite{SLBFGS}.

Our key observation of linear time Hessian-vector product computations for machine learning applications provides evidence that in such instances, obtaining true Hessian information is efficient enough to alleviate the need for quasi-Newton information via gradients.

\subsection{Discussion and Subsequent Work}
\label{sec:discussion}

In this section we provide a brief survey of certain technical aspects of our bounds which have since been improved by subsequent work.

An immediate improvement in terms of $S_2 \sim \kappa$ (in fact suggested in the original manuscript) was achieved by \cite{bollapragada2016exact} via conjugate gradient on a sub-sampled Hessian which reduces this to $\sqrt{\kappa}$. A similar improvement can also be achieved in theory through the extensions of LiSSA proposed in the paper. As we show in Section \ref{sec:experiments}, the worse dependence on condition number has an effect on the running time when $\kappa$ is quite large.\footnote{Equivalently, $\lambda$ is small.} Accelerated first-order methods, such as APCG \cite{lin2014accelerated}, outperform LiSSA in this regime. To the best of our knowledge second-order stochastic methods have so far not exhibited an improvement in that regime experimentally. We believe a more practical version of LiSSA-Sample could lead to improvements in this regime, leaving this as future work. 

To the best of our knowledge the factor of $S_1 = \kappa^2$ that appears to reduce the variance of our estimator has yet not been improved despite it being $O(1)$ in our experiments. This is an interesting question to which partial answers have been provided in the analysis of \cite{ye2017unifying}.

Significant progress has been made in the space of inexact Newton methods based on matrix sketching techniques. We refer the reader to the works of \cite{newtonsketch,xu2016sub,cohensubspaceembeddings,luo2016efficient, ye2017unifying} and the references therein.

We would also like to comment on the presence of a warm start parameter $\frac{1}{\kappa M}$ in our proofs of Theorems \ref{thm:mainthm} and \ref{thm:main_quadratic_thm}. In our experiments the warm start we required would be quite small
(often a few steps of gradient descent would be sufficient) to make LiSSA converge. The warm start does not affect the asymptotic results proven in Theorems \ref{thm:mainthm} and \ref{thm:main_quadratic_thm} because getting to such a warm start is independent of $\epsilon$. However, improving this warm start, especially in the context of Theorem \ref{thm:main_quadratic_thm}, is left as interesting future work. 

On the complexity side, \cite{arjevani2016oracle} proved lower bounds on the best running times achievable by second-order methods. In particular, they show that to get the faster rates achieved by LiSSA-Sample, it is necessary to use a non-uniform sampling based method as employed by LiSSA-Sample. We would like to remark that in theory, up to logarithmic factors, the running time of LiSSA-Sample is still the best achieved so far in the setting $m >> d$. Some of the techniques and motivations from this work were also generalized by the authors to provide faster rates for a large family of non-convex optimization problems \cite{agarwal2016finding}.

\subsection{Organization of the Paper}

The paper is organized as follows: we first present the necessary definitions, notations and conventions adopted throughout the paper in Section \ref{sec:preliminaries}. We then describe our estimator for LiSSA, as well as state and prove the convergence guarantee for LiSSA in Section \ref{sec:mainLissa}. After presenting a generic procedure to couple first-order methods with Newton's method in \ref{sec:extensions}, we present LiSSA-Sample and the associated fast quadratic solver in Section \ref{sec:lissaquadms}. We then present our results regarding self-concordant functions in Section \ref{sec:ellipsoidcovermain}. Finally, we present an experimental evaluation of LiSSA in Section \ref{sec:experiments}.

\section{Preliminaries}
\label{sec:preliminaries}
We adopt the convention of denoting vectors and scalars in lowercase, matrices in uppercase, and vectors in boldface. We will use $\|\cdot\|$ without a subscript to denote the
$\ell_2$ norm for vectors and the spectral norm for matrices. Throughout the paper we denote $\x^* \defeq \argmin_{\x \in \K} f(\x)$. A convex function $f$ is defined to be $\alpha$-strongly convex and $\beta$-smooth if, for all $\x, \y$, \[ \nabla f(\x)^{\top}(\y - \x) + \frac{\beta}{2}\|\y - \x\|^2  \geq f(\y) - f(\x) \geq \nabla f(\x)^{\top}(\y - \x) + \frac{\alpha}{2}\|\y - \x\|^2. \] The following is a well known fact about the inverse of a matrix $A$ s.t. $\|A\| \leq 1$ and $A \succ 0$:

\begin{equation}
\label{eqn:taylorexpansion}
 A^{-1} = \sum_{i=0}^{\infty} (I - A)^i.
 \end{equation}

\subsection{Definition of Condition Numbers}
\label{sec:conditionnumberappendix}
We now define several measures for the condition number of a function $f$. The
differences between these notions are subtle and we use them to precisely
characterize the running time for our algorithms.\footnote{During initial reading we suggest the reader to skip the subtlety with these notions with the knowledge that they are all smaller than the pessimistic bound one can achieve by considering a value proportional to $O(\lambda^{-1})$, where $\lambda$ is the coefficient of the $\ell_2$ regularizer.}

For an $\alpha$-strongly
convex and $\beta$-smooth function $f$, the condition number of the function is defined as $\kappa(f) \defeq \frac{\beta}{\alpha}$, or $\kappa$ when the function is clear from the context. Note that by definition this corresponds to the following notion:
\[ \kappa \defeq \frac{\max_\x \lambda_{\max}(\hess f(\x))}{\min_\x \lambda_{\min}(\hess f(\x))}\enspace. \]
We define a slightly relaxed notion of condition number where the $max$ moves out of the fraction above. We refer to this notion as a \textit{local} condition number $\kappa_l$ as compared to the \textit{global} condition number $\kappa$ defined above: 
\[ \kappa_l \defeq \max_{\x} \frac{\lambda_{\max}(\nabla^2 f(\x))}{\lambda_{\min}(\nabla^2 f(\x))}\enspace. \]
It follows that $\kappa_l \leq \kappa$. The above notions are defined for any general function $f$, but in the case of functions of the form 
$f(\x) = \frac{1}{m}\sum\limits_{k=1}^m f_k(\x)$, a further distinction is made with respect to the component functions. We refer to such definitions of the condition number by $\hat{\kappa}$. In such cases one typically assumes the each component is bounded by $\beta_{max}(\x) \defeq \max\limits_{k} \lambda_{\max}(\hess f_k(\x))$. The running times of algorithms like SVRG depend on the following notion of condition number: 
\[ \hat{\kappa} = \frac{\max_{\x} \beta_{\max}(\x)}{\min_{\x}{ \lambda_{\min}(\hess f(\x))}}\enspace.\]
Similarly, we define a notion of local condition number for $\hat{\kappa}$, namely
\[\hat{\kappa}_l \defeq \max\limits_{\x} \frac{\beta_{\max}(\x)}{\lambda_{\min}(\nabla^2 f(\x))},\]
and it again follows that $\hat{\kappa_l} \leq \hat{\kappa}$.

For our (admittedly pessimistic) bounds on the variance we also need a per-component strong convexity bound $\alpha_{\min}(\x) \defeq \min\limits_{k} \lambda_{\min}(\hess f_k(\x))$. We can now define 
\[\hat{\kappa}_l^{\max} \defeq \max\limits_{\x} \frac{\beta_{\max}(\x)}{\alpha_{\min}(\x)}\enspace.\]

\textit{Assumptions}: In light of the previous definitions, we make the following assumptions about the given function $f(\x) = \frac{1}{m}\sum\limits_{k=1}^m f_k(\x)$ to make the analysis easier. We first assume that the regularization term has been divided equally and included in $f_k(\x)$. We further assume that each $\hess f_k(\x) \preceq I$.\footnote{The scaling is
without loss of generality, even when looking at additive errors, as this gets picked up in the log-term due to the linear convergence.} We also assume that $f$ is $\alpha$-strongly convex
 and $\beta$-smooth, $\hat{\kappa}_l$ is the associated
 local condition number and $\hess f$ has a Lipschitz constant
 bounded by $M$.  

We now collect key concepts and pre-existing results that we
use for our analysis in the rest of the paper. 

\textit{Matrix Concentration}: The following lemma is a standard concentration of measure result for sums of
independent matrices.\footnote{The theorem in the reference states the
inequality for the more nuanced bounded variance case. We only state the simpler
bounded spectral norm case which suffices for our purposes.} An excellent
reference for this material is by
\cite{Tropp}.

\begin{theorem}[Matrix Bernstein, \cite{Tropp}]
\label{thm:MatrixBernstein}
Consider a finite sequence $\{X_k\}$ of independent, random, Hermitian matrices
with dimension $d$. Assume that 
\[ \av [X_k] = 0 \text{  and  } \|X_k\| \leq R.\]
Define $Y = \sum_k X_k$. Then we have for all $t \geq 0$,
\[ \pr \left( \|Y\| \geq t \right) \leq d
\exp\left(\frac{-t^2}{4R^2}\right).\]
\end{theorem}

\textit{Accelerated SVRG}: The following theorem was proved by \cite{Catalyst2015}.

\begin{theorem}
\label{thm:accsvrg}
Given a function $f(\x) = \frac{1}{m}\sum_{k=1}^{m} f_k(\x)$ with condition number $\kappa$, the accelerated version of SVRG via Catalyst \cite{Catalyst2015} finds an $\epsilon$-approximate minimum with probability $1 - \delta$ in time
\[ \tilde{O}(md + \min(\sqrt{\kappa m}, \kappa)d)\log\left(\frac{1}{\epsilon}\right).\]
\end{theorem}

\textit{Sherman-Morrison Formula}: The following is a well-known expression for writing the inverse of rank one perturbations of matrices:

\[\left( A + \bv \bv^T \right)^{-1} = A^{-1} - \frac{A^{-1}\bv\bv^T A^{-1}}{1 + \bv^T A^{-1} \bv}\enspace.\]

\section{LiSSA: Linear (time) Stochastic Second-Order Algorithm }
\label{sec:mainLissa}
In this section, we provide an overview of LiSSA (Algorithm \ref{alg:stochasticnewton}) along with its main convergence results.
\subsection{Estimators for the Hessian Inverse}
\label{sec:estimators}
Based on a recursive reformulation of the Taylor expansion (Equation \ref{eqn:taylorexpansion}), we may describe an unbiased estimator of the Hessian. For a matrix $A$, define $A^{-1}_j$ as the first $j$ terms in the Taylor expansion, i.e., \[A^{-1}_j \defeq \sum_{i=0}^{j} (I - A)^i,  \;\;\text{or equivalently} \;\;A^{-1}_j \defeq I + (I-A)A^{-1}_{j-1}\enspace. \]
Note that $\lim_{j \rightarrow \infty} A^{-1}_j \rightarrow A^{-1}$. Using the above recursive formulation, we now describe an unbiased estimator of
$\hessinv f$ by first deriving an unbiased estimator $\hestinv f_j$ for $\hessinv f_j$.
\begin{definition}[Estimator]
\label{def:estimator2}
Given $j$ independent and unbiased samples $\{X_1 \ldots X_j\}$ of the Hessian
$\hess f$, define $\{\hestinv f_0, \ldots, \hestinv f_j\}$ recursively as follows:
\[ \hestinv f_0 = I \text{ and } \hestinv f_t = I + (I-X_t)\hestinv f_{t-1} \;\;\text{for}\;\;t=1,\dots,j.\]
\end{definition}

It can be readily seen that $\av[\hestinv f_j] = \hessinv f_j$, and so
$\av[\hestinv f_j] \rightarrow \hessinv f$ as $j \rightarrow \infty$, providing us with
an unbiased estimator in the limit.

\begin{remark}One can also define and analyze a simpler (non-recursive) estimator based on directly sampling terms from the series in Equation \eqref{eqn:taylorexpansion}. Theoretically, one can get similar guarantees for the estimator; however, empirically our proposed estimator exhibited better performance. 
\end{remark}

\subsection{Algorithm}

Our algorithm runs in two phases: in the first phase it runs any efficient first-order method FO for $T_1$ steps to shrink the function value to the regime where we can then show linear convergence for our algorithm. It then takes approximate Newton steps based on the estimator from Definition \ref{def:estimator2} in place of the true Hessian inverse. We use two parameters, $S_1$ and $S_2$, to define the Newton step. $S_1$
represents the number of unbiased estimators of the Hessian inverse we average
to get better concentration for our estimator, while $S_2$
represents the depth to which we capture the Taylor expansion. In the algorithm, we compute the
average Newton step directly, which can be computed in linear time as observed earlier, instead of estimating the Hessian inverse. 

\label{section:stochasticnewton}

\begin{algorithm}[tb]
\caption{\textbf{LiSSA: Linear (time) Stochastic Second-Order Algorithm}}
\label{alg:stochasticnewton}
\begin{algorithmic}
\STATE {\bfseries Input: $T$, $f(\x) = \frac{1}{m}\sum\limits_{k = 1}^m f_k(\x)$, $S_1$,
$S_2$, $T_1$}
\STATE{$\x_1 = FO(f(\x), T_1)$ }
\FOR{$t = 1$ to $T$}
\FOR{$i = 1$ to $S_1$}
\STATE{$X_{[i,0]} = \nabla f(\x_t)$}
\FOR{$j = 1$ to $S_2$}
\STATE{Sample $\hest f_{[i,j]}(\x_t)$ uniformly from $\{\nabla^2 f_k(\x_t) \ | \
k \in [m]\}$}
\STATE{$X_{[i, j]} = \nabla f(\x_t) + (I - \hest f_{[i,j]}(\x_t))X_{[i, j-1]}$} \label{alg:line:lissastep}
\ENDFOR
\ENDFOR
\STATE{$X_t = 1/S_1 \left( \sum_{i=1}^{S_1} X_{[i,S_2]} 
 \right)$ }
\STATE{$\x_{t+1} = \x_t - X_t$}
\ENDFOR

\STATE{\textbf{return} $\x_{T+1}$}

\end{algorithmic}
\end{algorithm}

\subsection{Main Theorem}

In this section we present our main theorem which analyzes the convergence properties of LiSSA. Define $FO(M, \hat{\kappa}_l)$ to be the total time required by a first-order algorithm to achieve accuracy $\frac{1}{4\hat{\kappa}_lM}$. 
\begin{theorem}
\label{thm:mainthm}
Consider Algorithm \ref{alg:stochasticnewton}, and set the parameters as follows: $T_1=FO(M, \hat{\kappa}_l)$, $S_1 = O(\left(\hat{\kappa}^{\max}_l\right)^2 \ln(\frac{d}{\delta}))$, $S_2 \geq
      2\hat{\kappa}_l \ln(4\hat{\kappa}_l).$
The following guarantee holds for every $t \geq T_1$ with probability $1 - \delta$, 
\[ \|\x_{t+1} - \x^{*}\| \leq \frac{\|\x_{t} - \x^{*}\|}{2}.\]
Moreover, we have that each step of the algorithm takes at most $\tilde{O}(md +
\left(\hat{\kappa}^{\max}_l\right)^2\hat{\kappa}_l d^2)$ time. Additionally if $f$ is GLM, then each step of the algorithm can be run in time $\tilde{O}(md +
\left(\hat{\kappa}^{\max}_l\right)^2 \hat{\kappa}_l d)$.
\end{theorem}

As an immediate corollary, we obtain the following:

\begin{corollary}
\label{cor:main_corollary} 
For a GLM function $f(\x)$, Algorithm \ref{alg:stochasticnewton} returns a point $\x_t$ such
that with probability at least $1-\delta$, $$ f(\x_t) \leq \min_{\x^*} f(\x^*) +
\varepsilon$$ in total time $\tilde{O}(m +
\left(\hat{\kappa}^{\max}_l\right)^2\hat{\kappa}_l)d \ln\left(\frac{1}{
\varepsilon}\right) $ for $\epsilon \rightarrow 0$.
\end{corollary}

In the above theorems $\tilde{O}$ hides $\log$ factors of $\kappa, d,
\frac{1}{\delta}$. We note that our bound $(\hat{\kappa}^{\max}_l)^2$ on the variance is possibly pessimistic and can likely be improved to a more average quantity. However, since in our experiments setting the parameter $S_1 \sim O(1)$ suffices, we have not tried to optimize it further.

\label{sec:proofofmainlissatheorem}

We now prove our main theorem about the convergence of LiSSA (Theorem \ref{thm:mainthm}).
\begin{proof}[Proof of Theorem \ref{thm:mainthm}]

Note that since we use a first-order algorithm to get a solution of accuracy at least $\frac{1}{4 \hat{\kappa}_l M}$, we have that 
\begin{equation} 
\label{eqn:interimeqn}
\|\x_{1} - \x^{*}\| \leq \frac{1}{4 \hat{\kappa}_l M}.
\end{equation}

As can be seen from Definition \ref{def:estimator2}, a single step of our
algorithm is equivalent to $\x_{t+1} = \x_{t} -
\tilde{\nabla}^{-2}f(\x_t)\nabla f(\x_t)$, where $\tilde{\nabla}^{-2}f(\x_t)$ is the
average of $S_1$ independent estimators $\tilde{\nabla}^{-2}f(\x_t)_{S_2}$. We
now make use of the following lemma.

\begin{lemma}
\label{lemma:main_lemma}
Let $\x_{t+1} = \x_{t} -
\tilde{\nabla}^{-2}f(\x_t)\nabla f(\x_t)$, as per a single iteration of
Algorithm \ref{alg:stochasticnewton}, and suppose $S_1, S_2$ are as defined in
Algorithm \ref{alg:stochasticnewton}. Then if we choose $S_2 \geq 2\hat{\kappa}_l
\ln(2\hat{\kappa}_l)$ we have the following guarantee on the convergence rate for every
step with probability $1-\delta$: \[ \|\x_{t+1} - \x^*\| \leq \gamma \|\x_{t} - \x^*\| + M\|\hessinv f(\x_t)\|\|\x_t - \x^*\|^2 \] where $\gamma =
16\hat{\kappa}^{\max}_l \sqrt{\frac{\ln(d\delta^{-1})}{S_1}} + \frac{1}{16}$.
\end{lemma}

Substituting the values of $S_1$ and $S_2$, combining Equation
\eqref{eqn:interimeqn} and Lemma \ref{lemma:main_lemma}, and noting that
$\|\hessinv f(\x_t)\| \leq \hat{\kappa}_l$, we have that at the start of the
Newton phase the following inequality holds: \[\|\x_{t+1} - \x^*\| \leq  \frac{\|\x_{t}
- \x^*\|}{4} + M \hat{\kappa}^{\max}_l \|\x_t - \x^*\|^2 \leq \frac{\|\x_{t}
- \x^*\|}{2}\enspace.\]
It can be shown via induction that the above property holds for all $t
\geq T_1$, which concludes the proof.
\end{proof}

We now provide a proof of Lemma \ref{lemma:main_lemma}.
\begin{proof} [Proof of Lemma \ref{lemma:main_lemma}]  Define $\chi(\x_t) =
\int_0^1  \hess f(\x^* + \tau(\x_t - \x^*)) d\tau$. Note that $\nabla f(\x_t) =
\chi(\x_t) (\x_t - \x^{*})$. Following an analysis similar to that of
\cite{NesterovBook}, we have that
\begin{eqnarray*}
\|\x_{t+1} - \x^*\| &=& \|\x_{t} - \x^* - \hestinv f(\x_t)\grad
f(\x_t) \| \\
&=& \| \x_t - \x^* -  \hestinv f(\x_t) \chi(\x_t)(\x_t - \x^*)\|\\ 
&\leq& \| I -
\hestinv f(\x_t) \chi(\x_t)
\| \|\x_t - \x^*\|.
\end{eqnarray*}
Following from the previous equations, we have that 
\begin{equation*}
\frac{\|\x_{t+1} - \x^*\|}{\|\x_t - \x^*\|} \leq \| I -
\hestinv f(\x_t) \chi(\x_t)
\| = \| \underbrace{I - \hessinv f(\x_t)\chi(\x_t)}_{a} -
\underbrace{\left( \hestinv f(\x_t) - \hessinv f(\x_t)\right) 
\chi(\x_t)}_{b} \|.
\end{equation*}
We now analyze the above two terms $a$ and $b$ separately: 
\begin{eqnarray*}
\|a\| &=& \| I -  \hessinv f(\x_t) \chi(\x_t) \| \\
&\leq& \|\hessinv f(\x_t) \int_0^1 \left(\hess f(\x_t) - 
\hess f(\x^* + \tau(\x_t - \x^*)) d\tau \right)\|\\
&\leq& M \|\hessinv f(\x_t)\|  \| \x_t -
\x^* \|.
\end{eqnarray*}
The second inequality follows from the Lipschitz bound on the Hessian. The
second term can be bounded as follows:
\[ \|b\| \leq \left(
\| \left( \hestinv f(\x_t) - \hessinv f(\x_t) \right) \| \| \chi(\x_t) \| \right)
\leq \gamma. \]
The previous claim follows from Lemma \ref{lemma:concentration} which shows a
concentration bound on the sampled estimator and by noting that due to our
assumption on the function, we have that for all $\x,\;\|\hess f(\x)\| \leq 1$, and
hence $\|\chi(\x)\| \leq 1$.

Putting the above two bounds together and using the triangle inequality, we have that
\[\frac{\|\x_{t+1} - \x^*\|}{\|\x_t - \x^*\|} \leq M \|\hessinv f(\x_t)\|  \| \x_t -
\x^* \| + \gamma\]
which concludes the proof. 
 \end{proof}

\begin{lemma}
\label{lemma:concentration}
Let $\hestinv f(\x_t)$ be the average of $S_1$ independent samples of $\hestinv
f(\x_t)_{S_2}$, as defined in Definition \ref{def:estimator2} and used in the per-step update
of Algorithm \ref{alg:stochasticnewton}, and let $\hess f(\x_t)$ be the true Hessian. If we set $S_2 \geq 2\hat{\kappa}_l \ln(\hat{\kappa}_l S_1)$, then we have that
\begin{equation*}
\pr \biggr( \|\hestinv f(\x_t) - \hessinv f(\x_t)\| > 
16\hat{\kappa}^{\max}_l\sqrt{\frac{\ln(\frac{d}{\delta})}{S_1}} + 1/16 \biggr)
\leq
\delta.
\end{equation*}
\end{lemma}
\begin{proof}[Proof of Lemma \ref{lemma:concentration}]
\label{proof:lemBernstein}
First note the following statement which is a straightforward implication of our
construction of the estimator:
\[ \av[ \hestinv f(\x_t) ] = \av[ \hestinv f(\x_t)_{S_2} ] = \sum_{i =
0}^{S_2} (I - \hess f(\x_t))^i. \] We also know from Equation \eqref{eqn:taylorexpansion} that for
matrices $X$ such that $\|X\| \leq 1$ and $X \succ 0$,
\begin{equation*}
X^{-1} = \sum_{i = 0}^{\infty} (I - X)^i.
\end{equation*}
Since we have scaled the function such that $\|\hess f_k\| \leq 1$, it follows that \begin{equation} 
\label{eqn:sumseparation}
\hessinv f(\x_t) = \av\left[ \hestinv f(\x_t)_{S_2} \right] + \sum_{i
= S_2+1}^{\infty} (I - \hess f(\x_t))^i.
\end{equation}
Also note that since $\hess f(\x_t) \succeq \frac{I}{\hat{\kappa}_l}$, it follows
that $\| I -
\hess f(\x_t) \| \leq 1 - \frac{1}{\hat{\kappa}_l}$.
Observing the second term in the above equation,
\begin{eqnarray*} \|\sum_{i = S_2}^{\infty} (I -
\hess f(\x_t))^i\| &\leq& \|(I -
\hess f(\x_t))\|^{S_2}\left(\sum_{i = 0}^{\infty} \|I - \hess
f(\x_t)\|^i\right)  \\
&\leq& (1 - \frac{1}{\hat{\kappa}_l})^{S_2}\left( \sum_{i = 0}^{\infty} (1 -
\frac{1}{\hat{\kappa}_l})^i \right)
\\
&\leq& (1 - \frac{1}{\hat{\kappa}_l})^{S_2} \hat{\kappa}_l \\
&\leq& \exp\left(-\frac{S_2}{\hat{\kappa}_l}\right)\hat{\kappa}_l.
\end{eqnarray*}
Since we have chosen $S_2 \geq 2\hat{\kappa}_l \ln(4 \hat{\kappa}_l)$, we get that
the above term is bounded by $\frac{1}{16}$. 
We will now show, using the matrix Bernstein inequality (Theorem \ref{thm:MatrixBernstein}),
that the estimate $\hestinv f$ is concentrated around its expectation. To apply the inequality we first need to bound the spectral norm of
each random variable. To that end we note that $\hestinv f_{S_2}$ has
maximum spectral norm bounded by 
\[ \|\hestinv f_{S_2}\| \leq \sum_{i=0}^{S_2} (1 - 1/\hat{\kappa}^{\max}_l )^i \leq
\hat{\kappa}^{\max}_l.\] We can now apply Theorem \ref{thm:MatrixBernstein}, which gives the following: \begin{equation*}\pr \left( \|\hestinv f - \av [\hestinv f]\| > \epsilon \right)
\leq \\ 2d\;\exp \left(\frac{ - \epsilon^2 S_1}{64 (\hat{\kappa}^{\max}_l)^2}
\right).
\end{equation*}
Setting $\epsilon = 16\hat{\kappa}^{\max}_l\sqrt{\frac{ln(\frac{d}{\delta})}{S_1}}$
gives us that the probability above is bounded by $\delta$. Now putting together
the bounds and Equation \eqref{eqn:sumseparation} we get the required result.
\end{proof}

\subsection{Leveraging Sparsity}

A key property of real-world data sets is that although the input is a high
dimensional vector, the number of non-zero entries is typically very low. The following theorem shows that LiSSA can be implemented in a way to leverage the underlying
sparsity of the data. Our key observation is that for GLM functions, the rank one Hessian-vector product can be performed in $O(s)$ time where $s$ is the
sparsity of the input $\x_k$.

\begin{theorem} 
\label{thm:sparsitytheorem}
For GLM functions Algorithm \ref{alg:stochasticnewton} returns a point $\x_t$ such
that with probability at least $1-\delta$ $$ f(\x_t) \leq \min_{\x^*} f(\x^*) +
\varepsilon$$ in total time $\tilde{O}(ms + (\hat{\kappa}^{\max}_l)^2\hat{\kappa}_l
s)\ln\left(\frac{1}{ \varepsilon}\right) $ for $\epsilon \rightarrow 0$.
\end{theorem}

We will prove the following theorem, from which Theorem \ref{thm:sparsitytheorem} will immediately follow.

\begin{theorem}
\label{thm:sparsethm}
Consider Algorithm \ref{alg:stochasticnewton}, let $f$ be of the form described
above, and let $s$ be such that the number of non zero entries in $\x_i$ is
bounded by $s$. Then each step of the algorithm can be implemented in time $O(ms
+ (\kappa^{\max}_l)^2 \kappa_l s)$.
\end{theorem}

\begin{proof}[Proof of Theorem \ref{thm:sparsethm}]
Proof by induction. Let $c_0 = 1$, $d_0 = 1$, $\bv_0 = \mathbf{0}$, and consider the update rules $c_{j+1} = 1 + (1-\lambda)c_{j}$, $d_{j+1} = (1-\lambda)d_j$, and $\bv_{j+1} = \bv_{j} - \frac{1}{(1-\lambda)d_j}\hest f_{[i,j+1]}(\x)(c_j \nabla f(\x) + d_j\bv_j)$, where $\lambda$ is the regularization parameter.
For the base case, note that $X_{[i,0]} = c_0\nabla f(\x) + d_0\bv_0 = \nabla f(\x)$, as is the case in Algorithm \ref{alg:stochasticnewton}. Furthermore, suppose $X_{[i,j]} = c_j \nabla f(\x) + d_j\bv_j$. Then we see that 
\begin{align*}
X_{[i,j+1]} &= \nabla f(\x) + (I -\lambda I - \hest f_{[i,j+1]}(\x)) X_{[i,j]}\\
 &= \nabla f(\x) + ((1 - \lambda)I - \hest f_{[i,j+1]}(\x))(c_j \nabla f(\x) + d_j\bv_j)\\
 &= (1+(1-\lambda)c_j)\nabla f(\x) + (1-\lambda)(d_j\bv_j) - \hest f_{[i,j+1]}(\x)(c_j \nabla f(\x) + d_j\bv_j)\\
 &= c_{j+1}\nabla f(\x) + d_{j+1}\bv_{j+1}.
\end{align*}

Note that updating $c_{j+1}$ and $d_{j+1}$ takes constant time, and $\hest f_{[i,j+1]}(\x)(c_j
\nabla f(\x))$ and $\hest f_{[i,j+1]}(\x)\bv_j$ can each be calculated in $O(s)$
time. It can also be seen that each product gives an $s$-sparse vector, so
computing $\bv_{j+1}$ takes time $O(s)$. Since $\nabla f(\x)$ can be calculated in $O(ms)$ time, and since $\bv_0$ is 0-sparse which implies the number of non-zero entries of $\bv_j$ is at most $js$, it follows that the total time to calculate $X_t$ is $O(ms + (\kappa^{\max}_l)^2 \kappa_l s)$.
\end{proof}

\section{LiSSA: Extensions}
\label{sec:extensions}
In this section we first describe a family of algorithms which generically couple first-order methods as sub-routines with second-order methods. In particular, we formally describe the algorithm LiSSA-Quad (Algorithm \ref{alg:SVRGLiSSA}) and provide its runtime guarantee (Theorem \ref{thm:generalLiSSA}). The key idea underlying this algorithm is that Newton's method essentially reduces a convex optimization problem to solving intermediate quadratic sub-problems given by the second-order Taylor approximation at a point, i.e., the sub-problem $Q_t$ given by
\[ Q_t(\y) = f(\x_{t-1}) + \nabla f(\x_{t-1})^T \y + \frac{\y^T \hess f(\x_{t-1}) \y}{2}\] 
where $\y \defeq \x - \x_{t-1}$.
The above ideas provide an alternative implementation of our estimator 
for $\hessinv f(\x)$ used in LiSSA. Consider running gradient descent on the above
quadratic $Q_t$, and let $\y^i_t$ be the $i^{th}$ step in this process.
By definition we have that
\[ \y^{i+1}_t = \y^i_t - \nabla Q_t(\y^i_t) = (I - \hess
f(\x_{t-1}))\y^i_t - \nabla f(\x_{t-1}). \]
It can be seen that the above expression corresponds exactly to the steps taken in LiSSA (Algorithm \ref{alg:SVRGLiSSA}, line \ref{alg:line:lissastep}), the
difference being that we use a sample of the Hessian instead of the true
Hessian. Therefore LiSSA can also be interpreted as doing a partial stochastic gradient descent on the quadratic $Q_t$. It is partial because we have a precise estimate of gradient of the function $f$ and a stochastic estimate for the Hessian. We note that this is essential for the linear convergence guarantees we show for
LiSSA.

The above interpretation indicates the possibility of using any first-order linearly
convergent scheme for approximating the minimizer of the quadratic $Q_t$. 
In particular, consider any algorithm $ALG$ that, given a convex quadratic function
$Q_t$ and an error value $\epsilon$, produces a point $\y$ such that 
\begin{equation}
\label{eqn:algcondition}
\|\y - \y^*_{t}\| \leq \epsilon
\end{equation} 
with probability at least $1 - \delta_{ALG}$, where $\y^*_{t} = \argmin Q_t$. Let the total time taken by the
algorithm $ALG$ to produce the point be $T_{ALG}(\epsilon, \delta_{ALG})$. For our applications we require $ALG$ to be linearly convergent, i.e. $T_{ALG}$ is proportional to $ \log(\frac{1}{\epsilon})$ with probability at least $1 - \delta_{ALG}$.

Given such an algorithm $ALG$, LiSSA-Quad, as described in Algorithm \ref{alg:SVRGLiSSA}, generically implements the above idea, modifying LiSSA by replacing the inner loop with the given algorithm $ALG$. The following is a meta-theorem about the convergence properties of LiSSA-Quad.  
 
\begin{algorithm}[tb]
\caption{\textbf{LiSSA-Quad}}
\label{alg:SVRGLiSSA}
\begin{algorithmic}
\STATE {\bfseries Input: $T$, $f(\x) = \sum\limits_{k = 1}^m f_k(\x)$,
$ALG$, $ALG_{params}$, $T_1$, $\epsilon$ }
\STATE{$\x_0$ = ALG ($f(\x), T_1 $) }
\FOR{$t = 1$ to $T$}
\STATE{$Q_t(\y) = \nabla f(\x_{t-1})^T \y + \frac{\y^T \hess f(\x_{t-1}) \y}{2}$}
\STATE{$\x_t = A(Q_t, \epsilon^2, A_{params})$}
\ENDFOR
\STATE{\textbf{return} $\x_{T}$}

\end{algorithmic}
\end{algorithm}

\begin{theorem}
\label{thm:generalLiSSA}
Given the function $f(\x) = \sum f_i(\x)$ which is $\alpha$-strongly convex, let $\x^*$ be the minimizer of the function and $\{\x_t\}$ be defined as in Algorithm \ref{alg:SVRGLiSSA}.
Suppose the
algorithm $ALG$ satisfies condition (\ref{eqn:algcondition}) with
probability $1 - \delta_{ALG}$ under the appropriate setting of parameters
$ALG_{params}$.  Set the
parameters in the algorithm as follows: $T_1 = T_{ALG}(1 / 4\alpha M)$, $T = \log\log(1/\epsilon)$, $\delta_{ALG} = \delta/T$,
where $\epsilon$ is the final error guarantee one wishes to achieve. 
Then we have that after $T$ steps, with probability at least $1 - \delta$,
 \[\min_{t = \{1 \ldots T\}}\|\x_t - \x^*\| \leq \epsilon.\]
In particular, LiSSA-Quad(ALG) produces a point $\x$ such that 
\[ \|\x - \x^*\| \leq \epsilon \] in total time $O(T_{ALG}(\epsilon, \delta_{ALG}) \log\log(1/\epsilon))$ with probability at least $1 -\delta$ for $\epsilon \rightarrow 0$.
\end{theorem}
Note that for GLM functions, the $\nabla Q_t(\y)$ at any point can be computed in time linear in $d$. In particular, a full gradient of $Q_t$ can be computed in time $O(md)$ and a stochastic gradient (corresponding to a stochastic estimate of the Hessian) in time $O(d)$. Therefore, a natural choice for the algorithm $ALG$ in the above are first-order algorithms which are linearly convergent, for example SVRG, SDCA, and Acc-SDCA. Choosing a first-order algorithm FO gives us a family of algorithms LiSSA-Quad(FO), each with running time comparable to the running time of the underlying FO, up to logarithmic factors. The following corollary summarizes the typical running time guarantees for LiSSA-Quad(FO) when FO is Acc-SVRG. 

\begin{corollary}
\label{cor:lissaAccSDCA}
Given a GLM function $f(\x)$, if $ALG$ is replaced by Acc-SVRG \cite{Catalyst2015}, then under a suitable setting of
parameters, LiSSA-Quad produces a point $\x$ such that 
\[f(\x) - f(\x^*) \leq \epsilon\]
with probability at least $1 - \delta$, in total time $\tilde{O}(m+
\min\left\{\sqrt{\hat{\kappa}_l m}, \hat{\kappa}_l\right\}
)d\log(1/\epsilon)\log\log({1}/{\epsilon})$.
\end{corollary}

Here the $\tilde{O}$ above hides logarithmic factors in
$\kappa, d, \delta$, but not in $\epsilon$. Note that the above running times depend upon the condition number $\hat{\kappa}_l$ which as described in Section \ref{sec:preliminaries} can potentially provide better dependence compared to its global counterpart. In practice this difference could lead to faster running time for LiSSA-Quad as compared to the underlying first-order algorithm FO. We now provide a proof for Theorem \ref{thm:generalLiSSA}.

\begin{proof}[Proof of Theorem \ref{thm:generalLiSSA}]
We run the algorithm $A$ to achieve accuracy
$\epsilon^2$ on each of the intermediate quadratic functions $Q_t$, and we set $\delta_A = \delta/T$ which implies via a union bound that for all $t \leq T$,
\begin{equation}
\label{eqn:FOguarantee}
\| \x_{t+1} - \x^*_t \| \leq \epsilon^2
\end{equation} with probability at least $1 - \delta$.

Assume that for all $t
< T$, $ \|\x_t - \x^*\| \geq \epsilon$ (otherwise the theorem is trivially
true). Using the analysis of Newton's method as before, we get that for all $t \leq
T$,
\begin{eqnarray*}
\|\x_{t+1} - \x^* \| &\leq& \|\x^*_t - \x^*\| + \|\x_{t+1} - \x^*_t\| \\
&\leq& \|\x_t -\hessinv f(\x_t)\nabla f(\x_t) - \x^*\| + \|\x_{t+1} - \x^*_t\|
\\
&\leq& \frac{M}{4\alpha}\|\x_t - \x^*\|^2 + \epsilon^2 \\
&\leq& \left(\frac{M}{4\alpha} + 1\right) \|\x_t - \x^*\|^2 
\end{eqnarray*}
where the second inequality follows from the analysis in the proof of
Theorem \ref{thm:mainthm} and Equation \eqref{eqn:FOguarantee}. We know that $\|\x_0 - \x_t\| \leq \sqrt{\frac{\alpha}{M}}$ from the initial run of the first-order algorithm $FO$.
Applying the above inductively and using the value of $T$ prescribed by the theorem statement, we get that 
$\|\x_T - \x^*\| \leq \epsilon$.
\end{proof}

\section{Runtime Improvement through Fast Quadratic Solvers}
\label{sec:lissaquadms}

The previous section establishes the reduction from general convex optimization to quadratic functions. In this section we show how we can leverage the fact that for quadratic functions the running time for accelerated first-order methods can be improved in the regime when $\kappa > m >> d$. In particular, we show the following theorem.

\begin{theorem}
\label{thm:main_quadratic_thm}
Given a vector $\bb \in \reals^d$ and a matrix $A = \sum A_i$ where each $A_i$ is of the form $A_i = \bv_i\bv_i^T + \lambda I $ for some $\bv_i \in \reals^d,\|\vv_i\| \leq 1$ and $\lambda \geq 0$ a fixed parameter, Algorithm \ref{alg:quadraticsolver} computes a vector $\tilde{\vv}$ such that $\|A^{-1}\bb -
\tilde{\vv}\| \leq \epsilon$ with probability at least $1 - \delta$ in total time \[\tilde{O}\left(md\log\left(\frac{1}{\epsilon}\right) +
 \left(d + \sqrt{\kappa_{sample}(A)d}\right)d\log^2 \left(\frac{1}{\epsilon} \right) \right).\]
$\tilde{O}()$ contains factors logarithmic in $m,d,\kappa_{sample}(A),\|b\|,\delta$.
\end{theorem}

 $\kappa_{sample}(A)$ is the condition number of an $O(d\log(d))$ sized sample of A and is formally defined in Equation \eqref{eqn:defnkappasample}. We can now use Algorithm \ref{alg:quadraticsolver} to compute an approximate Newton step by setting $A = \nabla^2 f(x)$ and $\bb = \nabla f(x)$. We therefore propose LiSSA-Sample to be a variant of LiSSA-Quad where Algorithm \ref{alg:quadraticsolver} is used as the subroutine ALG and any first-order algorithm can be used in the initial phase. The following theorem bounding the running time of LiSSA-Sample follows immediately from Theorem \ref{thm:generalLiSSA} and Theorem \ref{thm:main_quadratic_thm}. 

\begin{theorem}
\label{thm:LiSSAquadms}
Given a GLM function $f(\x) = \sum_i f_i(\x)$, let $\x^* = \argmin f(\x)$. LiSSA-Sample produces a point $\x$ such that 
\[ \|\x - \x^*\| \leq \epsilon \] 
with probability at least $1 - \delta$ in total time 
\[\tilde{O}\left(\left(md\log\left(\frac{1}{\epsilon}\right) + \left(d + \sqrt{\kappa_{sample}(f)d}\right)d\log^2 \left(\frac{1}{\epsilon} \right) \right) \log\log\left(\frac{1}{\epsilon}\right)\right).\]
 $\tilde{O}()$ contains factors logarithmic in $m,d,\kappa_{sample}(f),G,\delta$.
\end{theorem}

\subsection{Fast Quadratic Solver - Outline}
In this section we provide a short overview of Algorithm \ref{alg:quadraticsolver}. To simplify the discussion, lets consider the case when we have to compute $A^{-1}\bb$ for a $d \times d$ matrix $A$ given as $A = \sum_{i = 1}^{m} \vv_i\vv_i^T = VV^T$ where the $i^{th}$ column of $V$ is $\vv_i$. The computation can be recast as minimization of a convex quadratic function $Q(\y) = \frac{\y^T A \y}{2} + \bb^Ty$ and can be solved up to accuracy $\epsilon$ in total time $\left(m + \sqrt{\kappa(A)m}\right)d\log(1/\epsilon)$ as can be seen from Theorem \ref{thm:accsvrg} Algorithm \ref{alg:quadraticsolver} improves upon the running time bound in the case when $m > d$. In the following we provide a high level outline of the procedure which is formally described as Algorithm \ref{alg:quadraticsolver}.

\begin{itemize}
\item Given $A$ we will compute a low complexity constant spectral approximation $B$ of $A$. Specifically $B = \sum_{i = 1}^{O(d\log(d))} \bu_i\bu_i^T$ and $B \preceq A \preceq 2B$. This is achieved by techniques developed in matrix sampling/sketching literature, especially those of \cite{YinTatPaper}. The procedure requires solving a constant number of $O(d\log(d))$ sized linear systems, which we do via Accelerated SVRG. 
\item We use $B$ as a preconditioner and compute $BA^{-1}\bb$ by minimizing the quadratic \\
$\frac{\y^T AB^{-1} \y}{2} + \bb^T\y$. Note that this quadratic is well conditioned and can be minimized using gradient descent. In order to compute the gradient of the quadratic which is given by $AB^{-1}\y$, we again use Accelerated SVRG to solve a linear system in B.
\item  Finally, we compute $A^{-1}\bb = B^{-1}BA^{-1}\bb$ using Accelerated SVRG to solve a linear system in B.
\end{itemize}

In the rest of the section we formally describe the procedure outlined and provide the necessary definitions. One key nuance we must take into account is the fact that based on our assumption we have included the regularization term into the component functions. Due to this, the Hessian does not necessarily look like a sum of rank one matrices. Of course, one can decompose the identity matrix that appears due to the regularizer as a sum of rank one matrices. However, note that the procedure above requires that each of the sub-samples must have good condition number too in order to solve linear systems on them with Accelerated SVRG. Therefore, the sub-samples generated must look like sub-samples formed from the Hessians of component functions. For this purpose we extend the procedure and the definition for leverage scores described by \cite{YinTatPaper} to the case when the matrix is given as a sum of PSD matrices and not just rank one matrices. We reformulate and reprove the basic theorems proved by \cite{YinTatPaper} in this context. To maintain computational efficiency of the procedure, we then make use of the fact that each of the PSD matrices actually is a rank one matrix plus the Identity matrix. We now provide the necessary preliminaries for the description of the algorithm and its analysis.

\subsection{Preliminaries for Fast Quadratic Solver}

For all the definitions and preliminaries below assume we are given a $d \times d$ PSD matrix $A \defeq \sum_{i=1}^{m} A_i$ where $A_i$ are also PSD matrices. Let $A \cdot B \defeq Tr(B^TA)$ be the standard matrix dot product. Given two matrices $A$ and $B$ we say $B$ is a $\lambda$-spectral approximation of $A$ if $\frac{1}{\lambda}A \preceq B \preceq A$.   

\begin{definition}[Generalized Matrix Leverage Scores]
Define
  \begin{equation}
      \tau_i(A) \defeq A^+ \cdot A_i      
  \end{equation}
  \begin{equation}
      \tau_i^B(A) \defeq B^+ \cdot A_i.   
  \end{equation}
\end{definition}
Then we have the following facts:
\begin{fact}
\begin{equation}
\label{eqn:tauupperbound}
\sum_{i=1}^{n} \tau_i(A) = Tr\left(\sum A^+A_i\right) = Tr(A^+A) = rank(A) \leq d.
\end{equation}
\end{fact}

\begin{fact}
      If $B$ is a $\lambda$-spectral approximation of $A$, then $\tau_i(A) \leq \tau_i^B(A) \leq \lambda \tau_i(A)$.
\end{fact}

Given $A$ of the form $A = \sum A_i$, we define a sample of $A$ of size $r$ as the following. Consider a subset of indices of size $r$, $I = \{i_1 \ldots i_{r}\} \subseteq [m]$. For every such sample $I$ and given a weight vector $\bw \in \reals^r$,  we can associate the following matrix:
\begin{equation}
      Sample(\bw, I) \defeq \sum_{j \in I} w_j A_j.
\end{equation}
When the sample is unweighted, i.e., $\bw = \mathbf{1}$, we will simply denote the above as $Sample(I)$. 
We can now define $\kappa_{sample}(A,r)$ to be 
\begin{equation}
\label{eqn:defnkappasample}
      \kappa_{sample}(A, r) \defeq \max_{I : |I| \geq r} \kappa(Sample(r, I)).
\end{equation}
This is the worst-case condition number of an unweighted sub-sample of size $r$ of the matrix $A$. For our results we will be concerned with a $\Omega(d \log(d))$ sized sub-sample, i.e., the quantity $\kappa_{sample}(A,O(d \log(d)))$. We remind the reader that by definition $\kappa_{sample}$ for Hessians of the functions we are concerned with is always larger than $1/\lambda$, where $\lambda$ is the coefficient of the $\ell_2$ regularizer.

The following lemma is a generalization of the leverage score sampling lemma (Lemma 4, \cite{YinTatPaper}). The proof is very similar to the original proof by \cite{YinTatPaper} and is included in the Appendix for completeness.

\begin{lemma}[Spectral Approximation via Leverage Score Sampling]
\label{lemma:generalleveragescorelemma}
      Given an error parameter $0 < \epsilon < 1$, let $\bu$ be a vector of leverage score overestimates, i.e., $\tau_i(A) \leq \bu_i$, for all $i \in [m]$. Let $\alpha$ be a sampling rate parameter and let c be a fixed positive constant. For each matrix $A_i$, we define a sampling probability $p_i(\alpha) = \min\{1, \alpha \bu_i c \log d\}$. Let $I$ be a random sample of indices drawn from $[m]$ by sampling each index with probability $p_i(\alpha)$. Define the weight vector $\bw(\alpha)$ to be the vector such that $\bw(\alpha)_i = \frac{1}{p_i(\alpha)}$. By definition of weighted samples we have that 
      \[Sample(\bw(\alpha),I) = \sum_{i=1}^{m} \frac{1}{p_i(\alpha)} A_i \mathbf{1}_{x_i \sim p_i(\alpha)}(x_i = 1) \]
      where $x_i$ is a Bernoulli random variable with probability $p_i(\alpha)$. 

      If we set $\alpha = \epsilon^{-2}$, $S = Sample(\bw(\alpha),I)$ is formed by at most $\sum_i \min\{1, \alpha \bu_i c \log(d)\} \leq \alpha c \log(d) \|\bu\|_1$ entries in the above sum, and $\frac{1}{1 + \epsilon} S$ is a $\frac{1 + \epsilon}{1 - \epsilon}$ spectral approximation for $A$ with probability at least $1 - d^{-c/3}$.
\end{lemma}

The following theorem is an analogue of the key theorem regarding uniform sampling (Theorem 1, \cite{YinTatPaper}). The proof is identical to the original proof and is also included in the appendix for completeness. 

\begin{theorem} [Uniform Sampling]
\label{thm:tautildetheorem}
Given any $A = \sum_{i=1}^{m} A_i$ as defined above, let $S = \sum_{j=1}^{r} X_j$ be formed by uniformly sampling $r$ matrices $X_1 \ldots X_r \sim \{A_i\}$ without repetition. Define 
\[\tilde{\tau}^S_i(A) =
\left\{
      \begin{array}{ll}
            \tau_i^S(A)  & \mbox{if } \exists \;j \;\;s.t.\ X_j = A_i \\
            \tau_i^{S + A_i}(A) & \mbox{otherwise}
      \end{array}
\right.\enspace.\] 
Then $\tilde{\tau}^S_i(A) \geq \tau_i(A)$ for all $i$, and 
\[ \expect\left[\sum_{i=1}^{n} \tilde{\tau}^S_i(A)\right] \leq \frac{md}{r}. \] 
\end{theorem}

Unlike the case of rank one matrices as was done by \cite{YinTatPaper}, it is not immediately clear how to efficiently compute $\tilde{\tau}_i$ as we cannot directly apply the Sherman-Morrison formula. In our case though, since we have that each $A_i = \bv_i\bv_i^T + \lambda I$ we define slightly different estimates which are efficiently computable and prove that they work as well. 

\begin{theorem}
\label{thm:tauhattheorem}
Suppose we are given any $A = \sum_{i=1}^{m} A_i$ where each $A_i$ is of the form $A_i = \bv_i\bv_i^T + \lambda I$. Let $S = \sum_{j=1}^{m} X_j$ be formed by uniformly sampling $r$ matrices $X_1 \ldots X_r \sim \{A_i\}$ without repetition. Define 
\[\hat{\tau}^S_i(A) =
\left\{
      \begin{array}{ll}
            \bv_i^T S^{+} \bv_i^T + \frac{d}{r} & \mbox{if } \exists \; j \;\; s.t.\ X_j = A_i \\
            \frac{1}{1 + \frac{1}{\bv_i^T (S + \lambda I)^{+} \bv_i^T}} + \frac{d}{r}& \mbox{otherwise}
      \end{array}
\right.\enspace.\] 
Then $\hat{\tau}^S_i(A) \geq \tau_i(A)$ for all $i$, and 
\[ \expect\left[\sum_{i=1}^{n} \hat{\tau}^S_i(A)\right] \leq O\left(\frac{md}{r}\right). \] 
\end{theorem}

\label{sec:FQSProofs}{}
\begin{proof}[Proof of Theo{}rem \ref{thm:tauhattheorem}]
We will prove that 
\begin{equation}
\label{eqn:subproofeqn}
\frac{d}{m} \geq \hat{\tau}_i^S(A) - \tilde{\tau}_i^S(A) \geq 0.
\end{equation}
Consider the case when $\exists \; j \;\; s.t.\ X_j = A_i$. By definition of $\tilde{\tau}_i^S(A)$,
\begin{multline*}
      \tilde{\tau}_i^S(A) = \tau_i^S(A) = S^+ \cdot A_i = S^+ \cdot (\bv_i\bv_i^T + \lambda I) = \bv_i^TS^{+}\bv_i + \lambda S^+ \cdot \lambda I \\
      \leq \bv_i^T S^{+} \bv_i^T + (r \lambda I)^+ \cdot I  = \bv_i^T S^{+} \bv_i^T + \frac{d}{r} = \hat{\tau}_i^S(A).
\end{multline*}

The above follows by noting that $S \succeq r \lambda I$. It also follows from the definition that $\hat{\tau}^S(A)_i - \tilde{\tau}^S(A)_i \leq \frac{d}{r}$.

In the other case a similar inequality can be shown by noting via the Sherman-Morrison formula that 
\[ \bv_i^T (S + \lambda I + \bv_i \bv_i^T)^+ \bv_i = \bv_i^T \left ( (S + \lambda I)^+ - \frac{(S + \lambda I)^+ \bv_i\bv_i^T (S + \lambda I)^+}{ 1 + \bv_i^T (S + \lambda I)^+ \bv_i} \right) \bv_i =  \frac{1}{1 + \frac{1}{\bv_i^T (S + \lambda I)^{+} \bv_i^T}}.\] 
This proves Equation \eqref{eqn:subproofeqn}. A direct application of Theorem \ref{thm:tautildetheorem} now finishes the proof.
\end{proof}

\subsection{Algorithms}

In the following we formally state the two sub-procedures: Algorithm \ref{alg:quadraticsolver}, which solves the required system, and Algorithm \ref{alg:YintatAlgo}, which is the sampling routine for reducing the size of the system. 

\begin{algorithm}[h!]
\caption{\textbf{REPEATED HALVING}}
\label{alg:YintatAlgo}
\begin{algorithmic}[1]
\STATE {\bfseries Input: $A = \sum_{i=1}^{m} (\vv_i\vv_i^T + \lambda I)$}
\STATE {\bfseries Output: $B$ an $O(d\log(d))$ size weighted sample of $A$ and $B \preceq A \preceq 2B$}

\STATE{Take a uniformly random unweighted sample of size $\frac{m}{2}$ of $A$ to form $A'$}
\IF{$A'$ has size $> O(d\log(d))$}
\STATE{Recursively compute an $2$-spectral approximation $\tilde{A'}$ of $A'$}
\ENDIF
\STATE{Compute estimates $\gamma_i$ of generalized leverage scores $\{\hat{\tau}^{A'}_i(A)\}$ s.t. the following are satisfied\[ \gamma_i \geq \hat{\tau}^{A'}_i(A)\]
\[\sum \gamma_i \leq \sum 16 \hat{\tau}^{A'}_i(A) + 1 \]}
\STATE{Use these estimates to sample matrices from $A$ to form $B$ }

\end{algorithmic}
\end{algorithm}

We prove the following theorem regarding the above algorithm REPEATED HALVING (Algorithm \ref{alg:YintatAlgo}).

\begin{theorem} 
\label{theorem:yintattheorem}
      REPEATED HALVING (Algorithm \ref{alg:YintatAlgo}) produces an $\tilde{O}(d)$ sized sample $B$ such that $\frac{B}{2} \preceq A \preceq 2B$ and can be implemented in total time 

      \[\tilde{O}\left(md + d^2 + \sqrt{\kappa_{sample}(O(d \log(d)))d}\right).\] 
      
\end{theorem}
The $\tilde{O}$ in the above contains factors logarithmic in $m,d,\|A\|_F$. Note that the Frobenius norm of $A$ is bounded by $d\|A\|$ which is bounded in our case by $d$.

\begin{algorithm}
\caption{\textbf{Fast Quadratic Solver} (FQS)}
\label{alg:quadraticsolver}
\begin{algorithmic}[1]
\STATE {\bfseries Input: $A = \sum_{i=1}^{m} (\vv_i\vv_i^T + \lambda I)$, $\bb$, $\epsilon$}
\STATE {\bfseries Output : $\tilde{\vv}$ s.t. $\|A^{-1}\bb - \tilde{\vv}\| \leq \epsilon$}
\STATE { Compute $B$ s.t. $2B \succeq A \succeq B$ using REPEATED HALVING(Algorithm \ref{alg:YintatAlgo})}
\STATE { $Q(\y) = \frac{\y^T AB^{-1} \y}{2} + \bb^T\y$ }
\STATE { Compute $\hat{\y}$ such that $ \|\hat{\y} - \argmin Q(\y)\| \leq \frac{\epsilon}{4\|B^{-1}\|}$}
\STATE { Output $\tilde{\vv}$ such that $\|B^{-1}\hat{\y} - \tilde{\vv}\| \leq \epsilon/2$}  
\end{algorithmic}

\end{algorithm}

We first provide the proof of Theorem \ref{thm:main_quadratic_thm} using Theorem \ref{theorem:yintattheorem} and then provide the proof for Theorem \ref{theorem:yintattheorem}. For the purpose of clarity of discourse we hide the terms that appear due to the probabilistic part of the lemmas. We can take a union bound to bound the total probability of failure, and those terms show up in the log factor. 

\begin{proof}[Proof of Theorem \ref{thm:main_quadratic_thm}]
We will first prove correctness which follows from noting that $\argmin_\y Q(\y) = BA^{-1}b$. Therefore we have that
\[\epsilon/2 \geq \|B^{-1} (\hat{\y} - \argmin_\y Q(\y))\| = \|B^{-1}\hat{\y} - A^{-1}b\|.\]
A simple application of the triangle inequality proves that $\|A^{-1}b - \tilde{\bv}\| \leq \epsilon$.

\textit{Running Time Analysis}:
We now prove that the algorithm can be implemented efficiently. We first note the following two simple facts: since $B$ is a $2$-approximation of $A$, $\kappa(B) \leq 2\kappa(A)$; furthermore, the quadratic $Q(\y)$ is $1$-strongly convex and $2$-smooth. Next, we consider the running time of implementing step 5. For this purpose we will perform gradient descent over the quadratic $Q(\y)$. Note that the gradient of $Q(\y)$ is given by 
      \[ \nabla Q(\y) = AB^{-1}\y - \bb. \]
      Let $\y_0 = 0$, $\tilde{\epsilon} = \left(\frac{\epsilon}{4 \|B^{-1}\|} \right)^2$ and $G_Q$ be a bound on the gradient of the quadratic Q(\y). Let $\vv_t$ and $\y_{t+1}$ be such that
      \[\|B^{-1}\y_{t-1} - \bv_t \| \leq \frac{\tilde{\epsilon}}{100 \|A\|G_Q}\]
      \[ \y_{t+1} = \y_t - \eta \left( A \vv_t - \bb \right). \]
      Define $h_t = Q(\y_t) - \min_\y Q(\y)$. Using the standard analysis of gradient descent, we show that the following holds true for $t > 0$:
      \[ h_t \leq \max \{\tilde{\epsilon}, (0.9)^t h_0 \}. \]
      This follows directly from the gradient descent analysis which we outline below. To make the analysis easier, we define a \textit{true} gradient descent series:
      \[ \z_{t+1} = \y_t - \eta \nabla Q(\y_t).\]

      Note that

      \[ \| \z_{t+1} - \y_{t+1}\| = \| A \left( \vv_t - B^{-1}\y_t \right)\| \leq \frac{\tilde{\epsilon}}{10 G_Q}. \]
      We now have that
      \begin{eqnarray*}
            h_{t+1} - h_t &=& Q(\y_{t+1}) - Q(\y_t) \\
            &\leq& \innerprod{\nabla Q(\y_t)}{\y_{t+1} - \y_t }  + \frac{\beta}{2}\|\y_{t+1} - \y_t\|^2 \\
            &=& \innerprod{\nabla Q(\y_t)}{ \z_{t+1} - y_t} + \innerprod{\nabla Q(\y_t)}{ \y_{t+1} - \z_{t+1}} + \frac{\beta}{2}\|\z_{t+1} - \y_t + \y_{t+1} - \z_{t+1} \|^2 \\
            &\leq& -\eta \|\nabla Q(\y_t)\|^2 + \innerprod{\nabla Q(\y_t)}{ \y_{t+1} - \z_{t+1}} + \beta\eta^2\|\nabla Q(\y_t)\|^2 + \beta\|\y_{t+1} - \z_{t+1} \|^2 \\ 
            &\leq& \frac{1}{\beta}\| \nabla Q(\y_t)\|^2 + \frac{2}{10\beta}\left((\|\nabla Q(\y_t)\|+1)\tilde{\epsilon} \right) \\
            &\leq& -\frac{\alpha}{\beta} h_t + \frac{\tilde{\epsilon}}{50\beta} \\        
      \end{eqnarray*} 
      where $\beta \geq 1$ and $\alpha \leq 2$ are the strong convexity and smoothness parameters of $Q(\y)$. Therefore, we have that 
      \[ h_{t+1} \leq 0.75 h_t + 0.08 \tilde{\epsilon}.\]
      Using the inductive assumption that $\|h_t\| \leq \max \{\tilde{\epsilon}, (0.9)^t h_0 \}$, it follows easily that 
      \[ h_{t+1} \leq \max \{\tilde{\epsilon}, (0.9)^{t+1} h_0 \}. \]
      Using the above inequality, it follows that for $t \geq O(\log(\frac{1}{\tilde{\epsilon}}))$, we have that $h_t$ is less than or equal to $\tilde{\epsilon}$. By the $1$-strong convexity of $Q(\y)$, we have that if $h_t \leq \tilde{\epsilon}$, then 
      \[ \|\y_t - \argmin_\y Q(\y)\| \leq \sqrt{\tilde{\epsilon}} \leq \frac{\epsilon}{4 \|B\|^{-1}}.\]
      The running time of the above sub-procedure is bounded by the time to multiply a vector with $A$, which takes $O(md)$ time, and the time required to compute $\bv_t$, which involves solving a linear system in $B$ at each step. Finally, in step 6 we once again compute the solution of a linear system in $B$. Combining these we get that the total running time is 

      \[ \tilde{O}(md + LIN(B, \hat{\epsilon})) \log\left(\frac{1}{\tilde{\epsilon}}\right)\]
where $\hat{\epsilon} = \frac{\epsilon}{400 \|A\|\|B\|^{-1} G_Q} = \Omega \left(\frac{\epsilon}{\kappa(A) G_Q} \right)$. Now we can bound $LIN(B,\epsilon)$ by $\tilde{O}(d^2 + d\sqrt{\kappa(A)d})\log(1/\epsilon)$ by using Acc-SVRG to solve the linear system and by noting that $B$ is an $O(d\log(d))$ sized 2-approximation sample of $A$. The bound $G_Q$ can be taken as a bound on the gradient of the quadratic at the start of the procedure, as the norm of gradient only shrinks along the gradient descent path. It is therefore enough to take $G_Q \leq \|b\|$, which finishes the proof.
\end{proof}

\begin{proof}[Proof of Theorem \ref{theorem:yintattheorem}]

The correctness of the algorithm is immediate from Theorems \ref{thm:tauhattheorem} and \ref{lemma:generalleveragescorelemma}. The key challenge in the proof lies in proving that there is an efficient way to implement the algorithm, which we describe next. 

\textit{Runtime Analysis}:
To analyze the running time of the algorithm we consider the running time of the computation required in step 7 of the algorithm. We will show that there exists an efficient algorithm to compute numbers $\gamma_i$ such that 
\[ \gamma_i \geq \hat{\tau}^{A'}_i(A)\]
\[\sum \gamma_i \leq \sum 16 \hat{\tau}^{A'}_i(A) + 8k \epsilon \|A\|_F^2. \]
 
First note that by recursion, $\tilde{A'}$ is a 2-spectral approximation of $A'$, and so we have that 
\[\hat{\tau}^{A'}_i(A) \leq \hat{\tau}^{\tilde{A}'}_i(A) \leq 2 \hat{\tau}^{A'}_i(A). \]

We also know that $A'$ is an $O(d\log(d))$ sized weighted sub-sample of $A'$, and so it is of the form
\[\tilde{A'} = \sum_{i=1}^{O(d\log(d))} \bv_i\bv_i^T + \lambda I = \sum_{i = 1}^{O(d\log(d))} b_ib_i^T\] where such a decomposition can be achieved easily via decomposing the identity into canonical vectors. Therefore, any such $\tilde{A'}$ can be written as $BB^T$, where the columns of $B$ are $b_i$. 

To compute $\hat{\tau}^{\tilde{A}'}_i(A)$, we need to compute the value of $\bv_i^T (\tilde{A}')^+ \bv_i = \|B^T (\tilde{A}')^+ \bv_i\|^2_2$. Indeed, computing this directly for all vectors $v_i$ may be inefficient, but one can compute a constant approximation using the following procedure outlined by \cite{YinTatPaper}. 

To compute a constant approximation, we randomly sample a Gaussian matrix $G$ consisting of $k$ rows and compute instead the norms $\gamma'(i) = \|G B^T (\tilde{A}')^+ \bv_i\|_2^2$. By the Johnson-Lindenstrauss lemma, setting $k = O(\log(md))$, we have that with high probability,
\[ 1/2 \|B^T (\tilde{A}')^+ \bv_i\|^2_2 \leq \gamma'(i) \leq 2 \|B^T (\tilde{A}')^+ \bv_i\|^2_2.\]

Consider the following procedure for the computation of $\gamma'(i)$:
\begin{enumerate}
      \item Compute the matrix $G'= G B^T$ first which takes $\tilde{O}(kd^2)$ time.
      \item For each row $G_i'$ of $G'$ compute a vector $G_i''$ such that $\|G_i'' - G_i'\tilde{A}'^+\|^2 \leq \epsilon$. This takes a total time of $k LIN(\tilde{A}',\epsilon)$.
      \item For each $v_i$ compute $\gamma_i'' \defeq \sum_{j = 1}^{k} <G_j'',\bv_i>^2$. This takes a total time of $O(kmd)$.
\end{enumerate}

Here, $LIN(S, \epsilon)$ is the running time of solving a linear system in $S$ with error $\epsilon$. Therefore, substituting $k = O(\log(md))$, the total running time of the above algorithm is
\[ O(md + d^2 + LIN(\tilde{A}', \epsilon)).\]
It is now easy from the definitions of $\gamma_i''$ and $\gamma_i'$ to see that 
\[\gamma_i'' \in \left[\gamma_i' - k\epsilon\|\bv_i\|^2, \gamma_i' + k\epsilon\|\bv_i\|^2 \right].\]
Setting $\gamma_i = 4(\gamma_i'' + k\epsilon\|\bv_i\|^2)$, it follows from the definitions that 
\[ \gamma_i \geq \hat{\tau}^{A'}_i(A)\]
\[\sum \gamma_i \leq \sum 16 \hat{\tau}^{A'}_i(A) + 8k \epsilon \|A\|_F^2. \]

Now setting $\epsilon = \frac{1}{8k \|V\|_F^2}$ satisfies the required inequality for $\gamma_i$. This implies that when sampling from $\gamma_i$, we will have a $2$-approximation, and the number of matrices will be bounded by $O(d \log(d))$.

To solve the linear systems required above we use Accelerated SVRG (Theorem \ref{thm:accsvrg}) which ensures that the running time of $LIN(S, \epsilon) \leq \tilde{O}(d^2 + d\sqrt{\kappa(S) d})\log(\frac{1}{\epsilon})$. This follows by noting that in our case $S$ is a sum of $O(d\log(d))$ matrices. To bound the condition number of $S$, note that it is a $2$-approximation of some unweighted sample of the matrix $A$ of size greater than $O(d\log(d))$. Therefore, we have that $\kappa(S) \leq \kappa_{sample}(A)$. 

Putting the above arguments together we get that the total running time of the procedure is bounded by 
\[ \tilde{O}(md + (d^2 + d\sqrt{\kappa_{sample(A)}d})),\]
which proves the theorem.
\end{proof}

\section{Condition Number Independent Algorithms}

\label{sec:ellipsoidcovermain}

In this section we state our main result regarding self-concordant functions and provide the proof. The following is the definition of a self-concordant function:

\begin{definition}[Self-Concordant Functions]
Let $\K \subseteq \reals^n$ be a non-empty open convex set, and let $f : \K \mapsto \reals$ be a $C^3$ convex function. Then, $f$ is said to be \text{self-concordant} if
$$|\nabla^3 f(\x)[\bh, \bh, \bh]| \leq 2(\bh^{\top} \hess f(\x) \bh)^{3/2},$$
where we have
\begin{equation*}
\nabla^k f(\x)[\bh_1, \dots, \bh_k] \defeq \frac{\partial^k}{\partial
t_1\dots\partial t_k} |_{t_1=\dots=t_k} f(\x+t_1\bh_1 + \dots + t_k\bh_k).
\end{equation*}
\end{definition}

Our main theorem regarding self-concordant functions is as follows:

\begin{theorem} 
\label{thm:selfconctheorem}
Let $0 < r < 1$, let $\gamma \geq 1$, and let $\nu$ be a
constant depending on $\gamma,r$. Set $\eta = 10(1 - r)^{2}$, $S_1 = c_r =
\frac{50}{(1-r)^4}$, where $c_r$ is a constant depending on $r$, and
$T=\frac{f(\x_1) - f(\x^*)}{\nu}$. Then, after $t > T$ steps, the following linear convergence guarantee
holds between two full gradient steps for Algorithm
\ref{alg:ellipsoidcover}:
\[\av \left[f(w_{s}) -
f(w^*) \right]  \leq  \\ \frac{1}{2}\av \left[f(w_{s-1}) -
f(w^*)\right].\]
Moreover, the time complexity of each full gradient step is $O(md + c_rd^2)$,
where $c_r$ is a constant depending on $r$ (independent of the condition
number).
\end{theorem}

To describe the algorithm we first present the requisite definitions regarding self-concordant functions. 

\subsection{Self-Concordant Functions Preliminaries}
An excellent reference for this material is the
lecture notes on this subject by \cite{NemirovskiBook}.

A key object in the analysis of self-concordant functions is the notion of
a Dikin ellipsoid, which is the unit ball around a point $\x$ in the norm given by
the Hessian $\|\cdot\|_{\hess f(\x)}$ at the point. We will refer to this norm
as the \textit{local norm} around a point and denote it as $\|\cdot\|_{\x}$.
Formally, we have: \begin{definition}[Dikin ellipsoid]
The Dikin ellipsoid of radius $r$ centered at a point $\x$ is defined as 
\[W_r(\x) \defeq \{\y\ |\ \|\y - \x\|_{\hess f(\x)} \leq r\}.\] 
\end{definition}
One of the key properties of self-concordant functions is that
inside the Dikin ellipsoid, the function is well conditioned with respect to the local norm
at the center, and furthermore, the function is smooth. The following lemmas makes these notions formal, and the proofs of these lemmas can
be found in the lecture notes of \cite{NemirovskiBook}.
\begin{lemma} [\cite{NemirovskiBook}] For all $\h$ such that
$\|\h\|_{\x} < 1$ we have that \[ (1 - \|\h\|_{\x})^2 \hess f(\x) \preceq \hess
f(\x + \h) \preceq \frac{1}{(1 - \|\h\|_{\x})^2}\hess f(\x).\]
\end{lemma}

\begin{lemma}[\cite{NemirovskiBook}]
\label{lemma:smoothness}
For all $\h$ such that
$\|\h\|_{\x} < 1$ we have that 
\[f(\x+\h) \leq f(\x) + \langle \nabla f(\x), \h \rangle + \rho\left(
\|\h\|_{\x} \right)
\]
where $\rho(x) \defeq -\ln(1 - x) - x$.
\end{lemma}
Another key quantity which is used both as a potential function as well as a
dampening for the step size in the analysis of Newton's method in general is the
Newton decrement which is defined as $\lambda(\x) \defeq \|\nabla f(\x)\|_{\x}^{*} =
\sqrt{\nabla f(\x)^{\top} \hessinv f(\x) \nabla f(\x)}$. The following lemma
quantifies how $\lambda(\x)$ behaves as a potential by showing that once it
drops below 1, it ensures that the minimum of the function lies in the current Dikin
ellipsoid. This is the property which we use crucially in our analysis.
\begin{lemma} [\cite{NemirovskiBook}]
\label{lemma:lambdalemma}
If $\lambda(\x) < 1$ then 
\[ \|\x - \x^*\|_{\x}  \leq \frac{\lambda(\x)}{1 - \lambda(\x)}.\]
\end{lemma}

\subsection{Condition Number Independent Algorithms}

In this section we describe an efficient linearly convergent
method (Algorithm \ref{alg:ellipsoidcover}) for optimization of self-concordant
functions for which the running time is independent of the condition number. We have not
tried to optimize the complexity of the algorithms in terms of $d$
as our main focus is to make it condition number independent.

The key idea here is the ellipsoidal view of Newton's method, whereby we show that after making a constant number of full Newton steps, one can identify an ellipsoid and a norm such that the
function is well conditioned with respect to the norm in the ellipsoid. This
is depicted in Figure \ref{fig:ellipsoidPicture}. 

At this point one can run any desired first-order algorithm.
In particular, we choose SVRG and prove its fast convergence. Algorithm \ref{alg:modifiedSVRG} (described in the appendix) states the
modified SVRG routine for general norms used in Algorithm
\ref{alg:ellipsoidcover}.

\begin{figure}[t]
\vskip 0.2in

\begin{center}
\centerline{\includegraphics[width=\columnwidth]{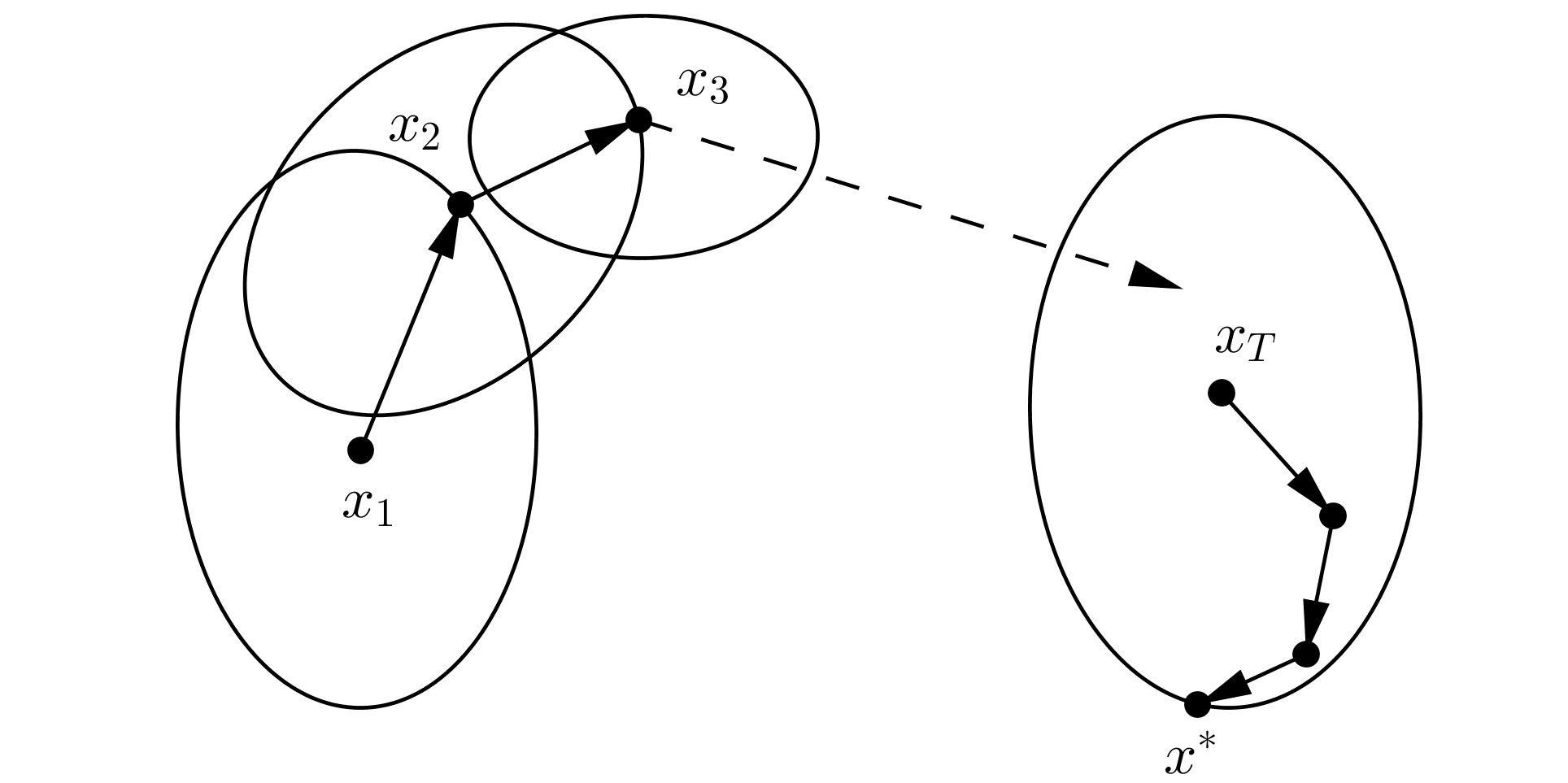}}
\caption{Visualization of the steps taken in Algorithm \ref{alg:ellipsoidcover}.}
\label{fig:ellipsoidPicture}
\end{center}
\vskip -0.2in
\end{figure}

\begin{algorithm}[b]
\caption{\textbf{ellipsoidCover}}
\label{alg:ellipsoidcover}
\begin{algorithmic}[1]
\STATE \textbf{Input :} Self-concordant $f(\x) = \sum\limits_{i=1}^m f_i(\x)$,
$T$, $r \in \{0,1\}$, initial point $\x_1 \in \K$, $\gamma > 1$, $S_1$,
$\eta$, $T$

\STATE{Initialize: $\x_{\mathrm{curr}} = \x_1$}
\WHILE{$\lambda(\x_{\mathrm{curr}}) > \frac{r}{1+r}$}
\STATE{$step = \gamma(1+\lambda(\x_{\mathrm{curr}}))$}
\STATE{$\x_{\mathrm{curr}} = \x_{\mathrm{curr}} -
\frac{1}{step} \hessinv f(\x_{\mathrm{curr}}) \grad f(\x_{\mathrm{curr}})$}
\ENDWHILE

\STATE{$\x_{T+1} = \mathbf{N-SVRG}(W_r(\x), f(\x), \hess
f(\x_{\mathrm{curr}}), S_1, \eta, T)$}

\STATE{\textbf{return} $\x_{T+1}$}

\end{algorithmic}
\end{algorithm}

We now state and prove the following theorem regarding the convergence of
Algorithm \ref{alg:ellipsoidcover}.

\begin{proof}
It follows from Lemma \ref{lemma:ConstantDecreaseLemma} that at $t =
\frac{f(\x_1) - f(\x^*)}{\nu}$, the minimizer is contained in the Dikin
ellipsoid of radius $r$, where $\nu$ is a constant depending on $\gamma,r$. This fact,
coupled with Lemma \ref{lemma:LocalConditioning}, shows that the function
satisfies the following property with respect to $W_{\x_t}$:
\[ \forall \x \in W_{\x_t}\;\;(1-r)^2 \hess f(\x_T) \preceq \hess f(\x) \preceq
(1-r)^{-2} \hess f(\x_T).\]
Using the above fact along with Lemma \ref{lemma:SVRGLemma}, and substituting for the
parameters, concludes the proof.
\end{proof}

We describe the Algorithm N-SVRG (Algorithm \ref{alg:modifiedSVRG} in the appendix). Since the algorithm and the following lemmas are minor variations of their original versions, we include the proofs of the following lemmas in the appendix for completeness. 

\begin{lemma}
\label{lemma:LocalConditioning}
Let $f$ be a self-concordant function over $\K$, let $0 < r < 1$, and consider
$\x \in \K$. Let $W_{r}(\x)$ be the Dikin ellipsoid of radius $r$, and let $\alpha = (1-r)^2$ and $\beta = (1-r)^{-2}$. Then, for all $\bh$ s.t. $\x + \bh \in W_{r}(\x)$, \[\alpha\hess f(\x) \preceq \hess f(\x + \h) \preceq \beta \hess f(\x).\]
\end{lemma}

\begin{lemma}
\label{lemma:ConstantDecreaseLemma}
Let $f$ be a self-concordant function over $\K$, let $0 < r < 1$, let $\gamma  \geq 1$, and consider following 
the damped Newton step as described in Algorithm \ref{alg:ellipsoidcover} with initial point $\x_1$. Then, the number of 
steps $t$ of the algorithm before the minimizer of $f$ is contained in the Dikin ellipsoid of radius $r$ of the current iterate, 
i.e. $\x^* \in W_{r}(\x_t)$, is at most $t = \frac{f(\x_1) - f(\x^*)}{\nu}$, where 
$\nu$ is a constant depending on $\gamma,r$.
\end{lemma}

\begin{lemma}
\label{lemma:SVRGLemma}
Let $f$ be a convex function. Suppose there exists a convex set $\K$ and a
positive semidefinite matrix $A$ such that for all $\x \in \K$, $\alpha A
\preceq \hess f(\x) \preceq \beta A $.
Then the following holds between two full gradient steps of Algorithm
\ref{alg:modifiedSVRG}:
\begin{equation*} \av \left[f(\x_{s}) -
f(\x^*) \right]  \leq  \left( \frac{1}{\alpha\eta (1 - 2\eta \beta)n} +
\frac{2\eta \beta}{(1 - 2\eta \beta)}\right) \av \left[f(\x_{s-1}) -
f(\x^*)\right].\end{equation*}
\end{lemma}

\section{Experiments}
\label{sec:experiments}

In this section we present experimental evaluation for our theoretical results.\footnote{Our code for LiSSA can be found here: \url{https://github.com/brianbullins/lissa_code}.} We perform the experiments for a classification task over two
labels using the logistic regression (LR) objective
function with the $\ell_2$ regularizer. For all of the classification tasks we choose two values of $\lambda$:$\frac{1}{m}$ and $\frac{10}{m}$, where $m$ is the number of training examples. We perform the above classification tasks over four data sets:
MNIST, CoverType, Mushrooms, and RealSIM. Figure \ref{fig:fourdatasets} displays the log-error achieved by LiSSA as compared to two standard first-order algorithms, SVRG and SAGA \cite{SVRG,SAGA}, in terms of the number of passes over the data. Figure \ref{fig:NewtoncomparisonTime} presents the performance of LiSSA as compared to NewSamp \cite{newsamp} and standard Newton's method with respect to both time and iterations.
\begin{figure}{}
\begin{center}
\setlength{\abovecaptionskip}{0pt minus 5 pt}
\includegraphics[width=1.0\textwidth]{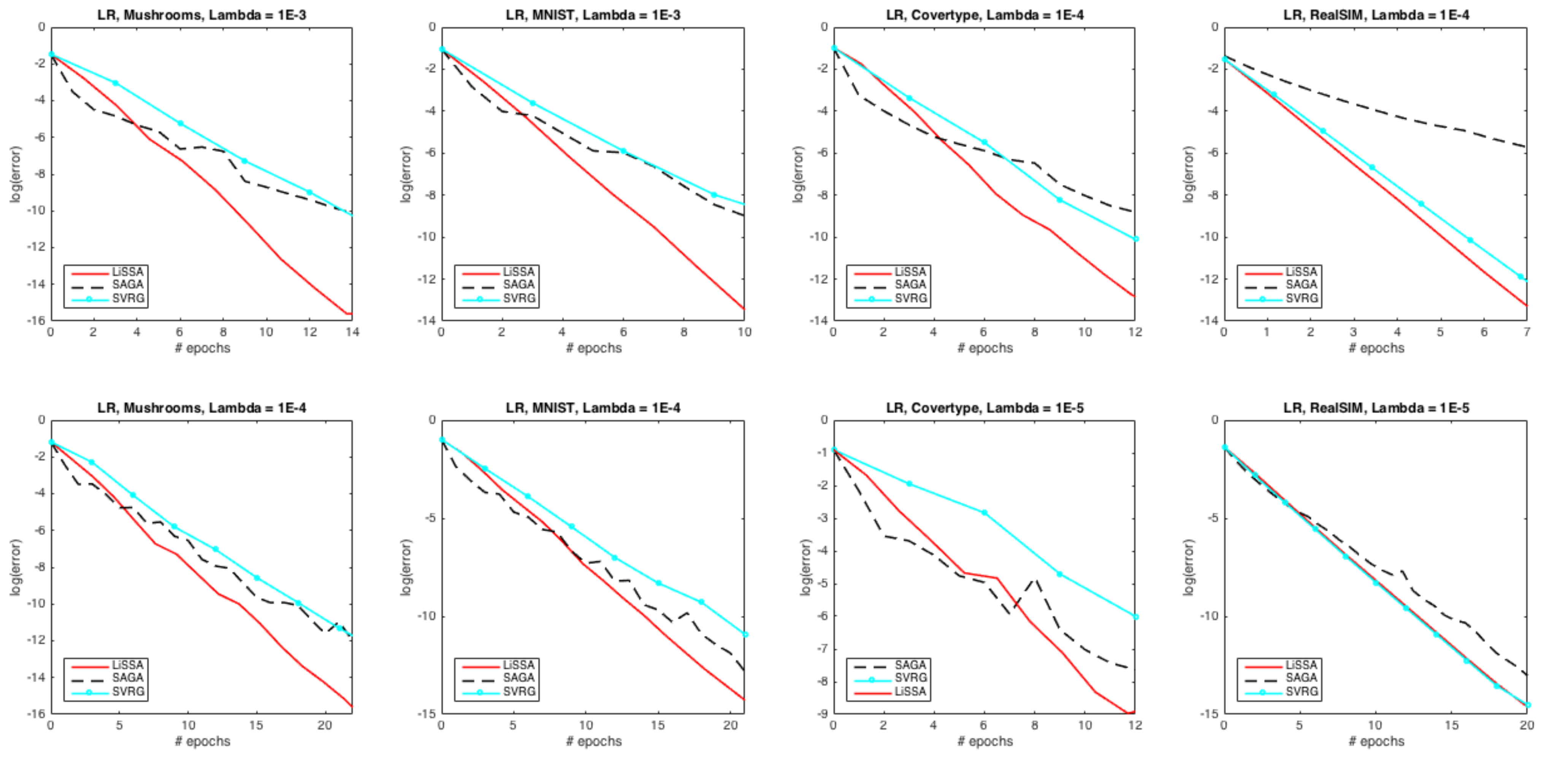}
\caption{Performance of LiSSA as compared to a variety of related optimization
methods for different data sets and choices of regularization
parameter $\lambda$. $S_1 = 1$, $S_2 \sim \kappa \ln(\kappa)$.}
\label{fig:fourdatasets}
\end{center}
\end{figure}

\begin{figure}
\begin{center}
\centerline{\includegraphics[width=0.8\columnwidth]{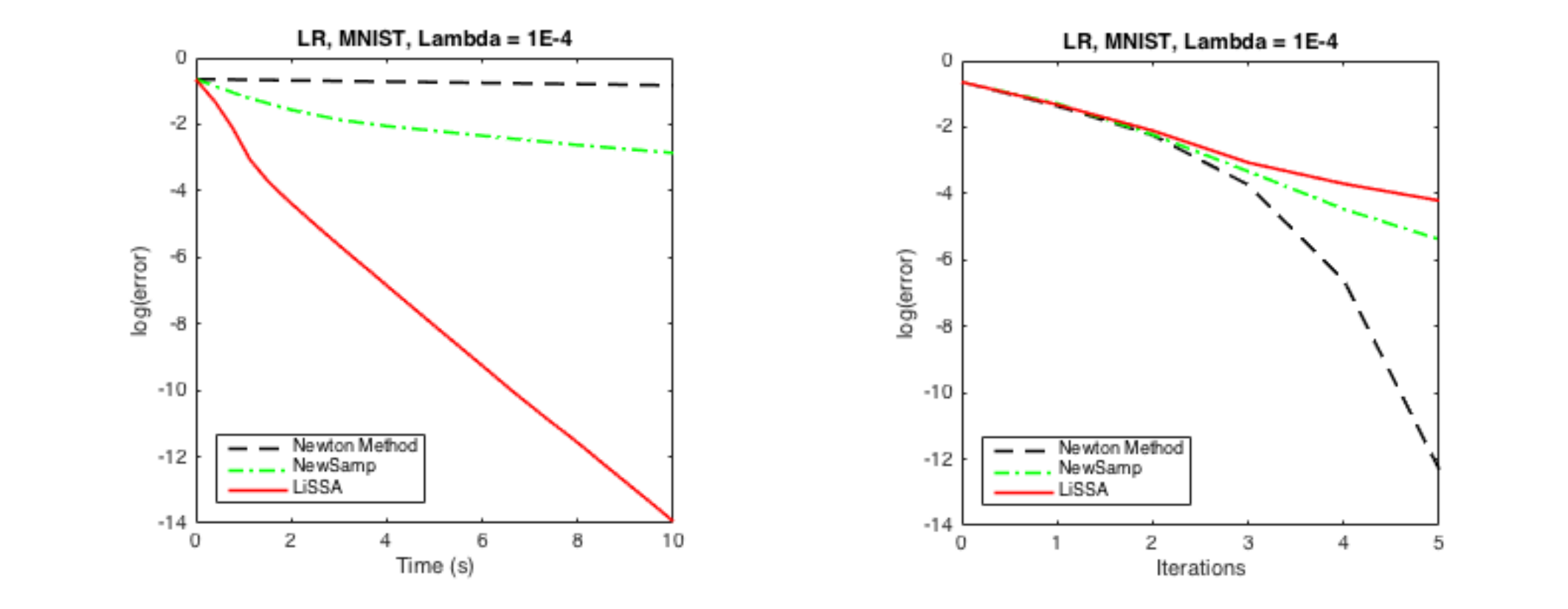}}
\caption{Convergence of LiSSA over time/iterations for logistic regression with MNIST, as
compared to NewSamp and Newton's method.}
\label{fig:NewtoncomparisonTime}
\end{center}

\vskip -0.2in
\end{figure}

\subsection{Experiment Details}
\label{sec:experimentsappendix}

In this section we describe our experiments and choice of parameters in detail. Table \ref{table:datasets} provides details regarding the data sets chosen for the experiments. To make sure our functions are scaled such that the norm of the Hessian is bounded, we scale the above data set points to unit norm.

\begin{table}[h!]
\caption{A description of data sets used in the experiments.}
\label{table:datasets}
\begin{center}
\begin{small}
\begin{sc}
\begin{tabular}{lcccr}
\hline

Data set & m & d & Reference \\
\hline
MNIST4-9     & 11791 & 784 & \cite{mnist}\\
Mushrooms    & 8124 & 112 & \cite{uci}\\
CoverType    & 100000 & 54  & \cite{covertype}\\
RealSIM    & 72309 & 20958  & \cite{realsim}\\
\hline
\end{tabular}
\end{sc}
\end{small}
\end{center}
\vskip -0.1in
\end{table}

 \subsection{Comparison with Standard Algorithms}

In Figures \ref{fig:fourdatasets} and \ref{fig:threedatasetstime} we present comparisons between the
efficiency of our algorithm with different standard and
popular algorithms. In both cases we plot $\log(CurrentValue - OptimumValue)$. We obtained the optimum value for each
case by running our algorithm for a long enough time until it converged to the point of machine
precision. 

\textit{Epoch Comparison}: In Figure \ref{fig:fourdatasets}, we compare LiSSA with SVRG and SAGA in terms of the accuracy achieved versus the number of passes over the data. To compute the number of passes in SVRG and SAGA, we make sure that the inner stochastic gradient iteration in both the algorithms counts as exactly one pass. This is done because although it accesses gradients at two different points, one of them can be stored from before in both cases. The outer full gradient in SVRG counts as one complete pass over the data. We set the number of inner iterations of SVRG to $2m$ for the case when $\lambda = 1/m$, and we parameter tune the number of inner iterations when $\lambda = 10/m$. The stepsizes for all of the algorithms are parameter tuned by an exhaustive search over the parameters. 

\textit{Time Comparison}: For the comparison with respect to time (Figure \ref{fig:threedatasetstime}), we consider the following algorithms: AdaGrad \cite{adagrad}, BFGS \cite{bro70, fle70, gol70, sha70}, gradient descent,
SGD, SVRG \cite{SVRG} and SAGA \cite{SAGA}. 
The $\log(Error)$ is plotted as a function of the time elapsed from the start of the run for each algorithm. We next describe our choice of parameters for the algorithms. 
For AdaGrad we used the faster diagonal scaling version proposed by \cite{adagrad}. 
We implemented the basic version of BFGS with backtracking line search. 
In each experiment for gradient descent, we find a reasonable step size using parameter tuning.
For stochastic gradient descent, we use the variable step size $\eta_t =
\gamma/\sqrt{t}$ which is usually the prescribed step size, and we hand tune the
parameter $\gamma$. The parameters for SVRG and SAGA were chosen in the same way as before.

\textit{Choice of Parameters for LiSSA}: To pick the parameters for our algorithm, we observe that it exhibits smooth behavior even in the case of $S_1 = 1$, so this is used for the experiments. However, we observe that increasing $S_2$ has a positive effect on the convergence of the algorithm up to a certain point, as a higher $S_2$ leads to a larger per-iteration cost. This behavior is consistent with the theoretical analysis. We summarize the
comparison between the per-iteration convergence and the value of $S_2$ in Figure
\ref{fig:S2comparison}. As the theory predicts $S_2$
to be of the order $\kappa\ln(\kappa)$, for our experiments we determine an
estimate for $\kappa$ and set $S_2$ to around $\kappa\ln(\kappa)$. This value
is typically equal to $m$ in our experiments. We observe that setting $S_2$
in this way resulted in the experimental results displayed in Figure
\ref{fig:fourdatasets}.

\textit{Comparison with Second-Order Methods}:
Here we present details about the comparison between LiSSA, NewSamp
\cite{newsamp}, and standard Newton's method, as displayed in Figure \ref{fig:NewtoncomparisonTime}. We perform this experiment on the
MNIST data set and show the convergence properties of all three algorithms over
time as well as over iterations. We could not replicate the results of NewSamp
on all of our data sets as it sometimes seems to diverge in our experiments.
For logistic regression on the MNIST data set, we could
get it to converge by setting the value of $S_1$ to be slightly higher. We observe as is predicted
by the theory that when compared in terms of the number of iterations, NewSamp
and LiSSA perform similarly, while Newton's method performs the best as it attains a
quadratic convergence rate. This can be seen in Figure
\ref{fig:NewtoncomparisonTime}. However, when we consider the performance in terms of time for these algorithms, we see that LiSSA has a significant
advantage.

\textit{Comparison with Accelerated First-Order Methods}:
Here we present experimental results comparing LiSSA with a popular accelerated first-order method, APCG \cite{lin2014accelerated}, as seen in Figure \ref{fig:APCGCompare}. We ran the experiment on the RealSIM data set with three settings of $\lambda = 10^{-5},\ 10^{-6},\ 10^{-7}$, to account for the high condition number setting. We observe a trend that can be expected from the runtime guarantees of the algorithms. When $\lambda$ is not too low, LiSSA performs better than APCG, but as $\lambda$ gets very low we see that APCG performs significantly better than LiSSA. This is not surprising when considering that the running time of APCG grows proportional to $\sqrt{\kappa m}$, whereas for LiSSA the running time can at best be proportional to $\kappa$. We note that for accelerated first-order methods to be useful, one needs the condition number to be quite large which is not often the case for applications. Nevertheless, we believe that an algorithm with running time guarantees similar to LiSSA-Sample can get significant gains in these settings, and we leave this exploration as future work. 
\begin{center}
\begin{figure}[h!]
\setlength{\abovecaptionskip}{0pt minus 5 pt}
\includegraphics[width=\textwidth]{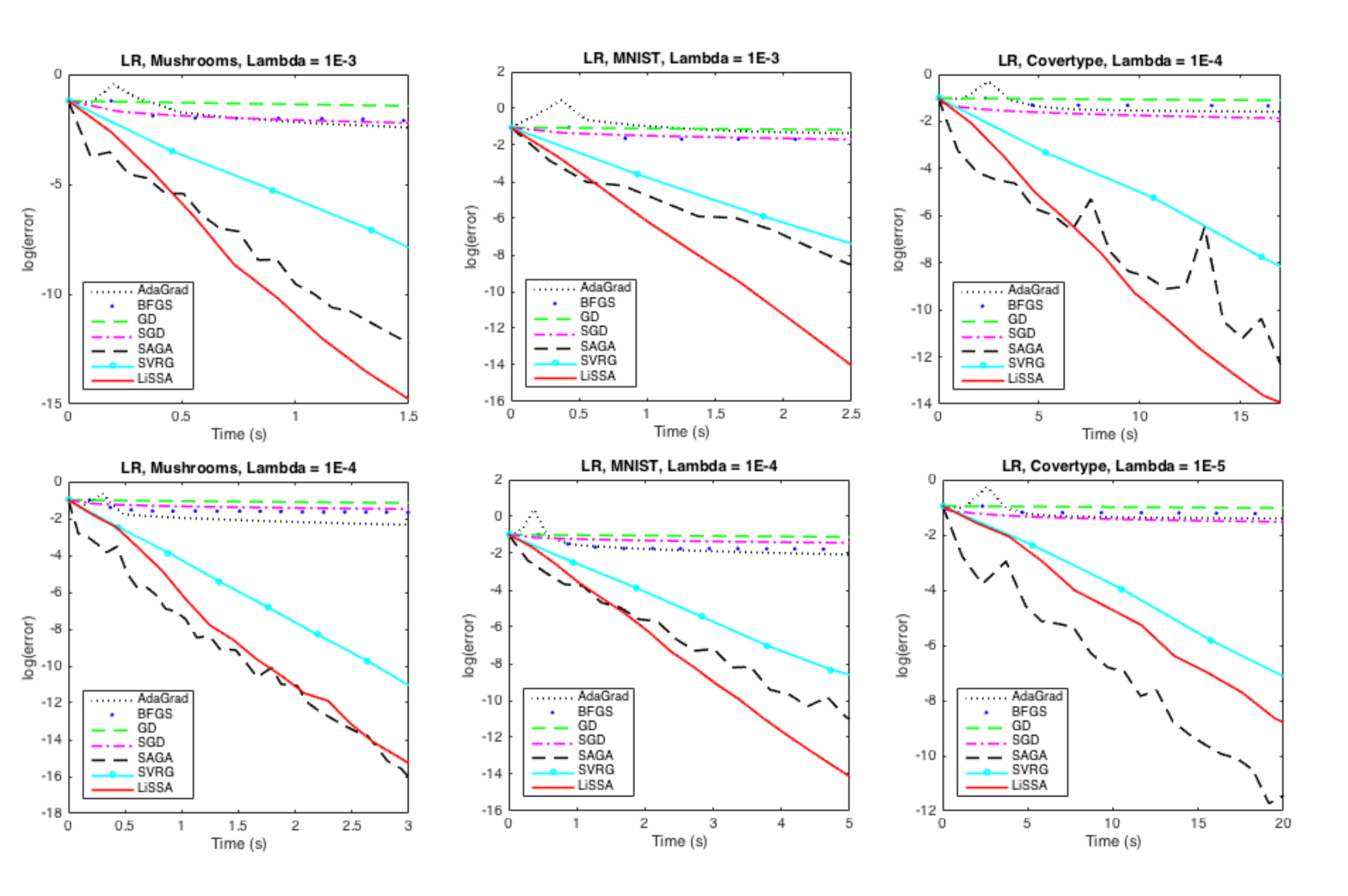}
\caption{Performance of LiSSA as compared to a variety of related optimization
methods for different data sets and choices of regularization
parameter $\lambda$. $S_1 = 1$, $S_2 \sim \kappa \ln(\kappa)$.}
\label{fig:threedatasetstime}
\end{figure}
\end{center}
\begin{figure}[h!]
\vskip 0.2in
\begin{center}
\centerline{\includegraphics[width=\textwidth]{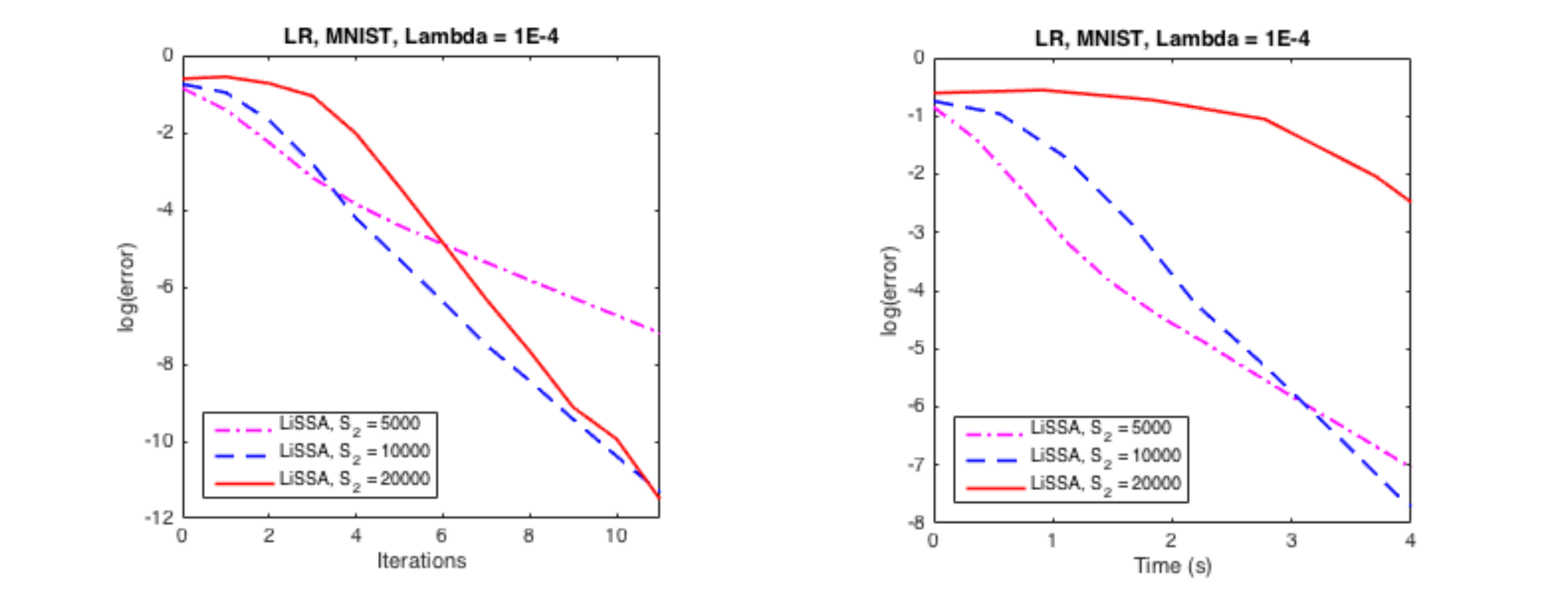}}
\caption{Differing convergence rates for LiSSA based on different choices of the $S_2$ parameter.}
\label{fig:S2comparison}
\end{center}
\vskip -0.2in
\end{figure} 
\begin{figure}[h!]
\vskip 0.2in
\begin{center}
\centerline{\includegraphics[width=\textwidth]{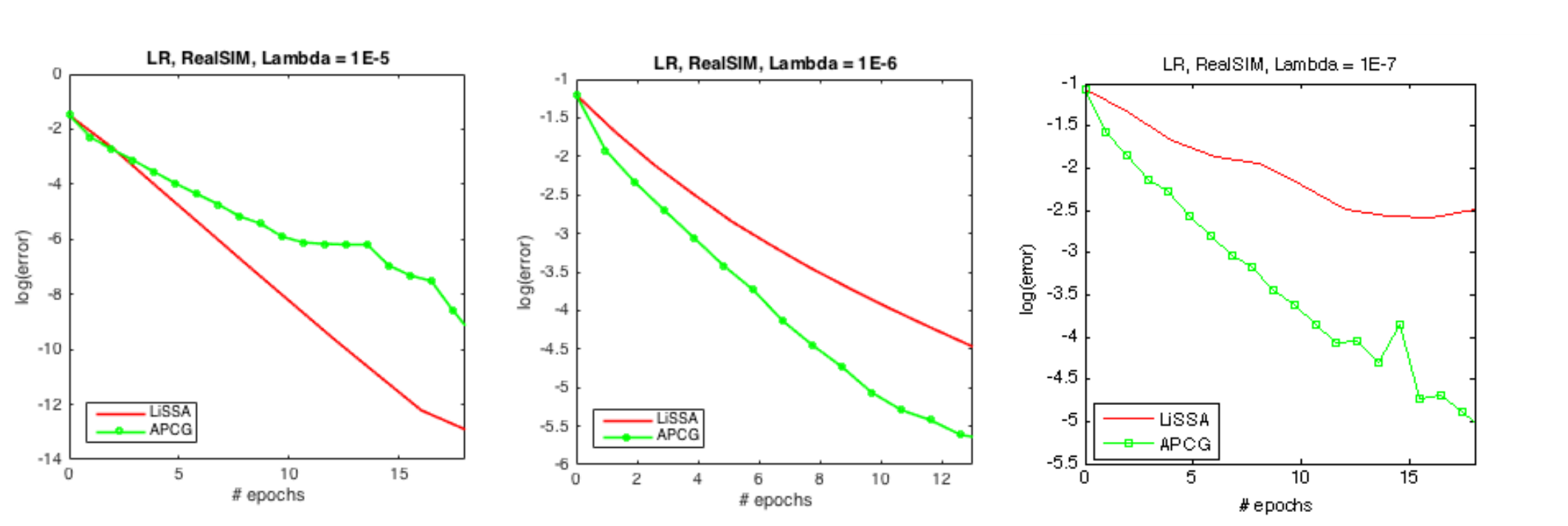}}
\caption{Comparison with accelerated first-order methods in ill-conditioned settings.}
\label{fig:APCGCompare}
\end{center}
\vskip -0.2in
\end{figure} 

\section{Acknowledgements}
The authors would like to thank Zeyuan Allen-Zhu, Dan Garber, Haipeng Luo and David McAllester for several illuminating discussions and suggestions. We would especially like to thank Aaron Sidford and Yin Tat Lee for discussions regarding the matrix sampling approach.

\newpage

\bibliographystyle{alpha}
\bibliography{references}

\appendix
\section{Remaining Proofs}
\label{app:theorem}
Here we provide proofs for the remaining lemmas.
\subsection{Proof of Lemma \ref{lemma:generalleveragescorelemma}}
\begin{proof}
The proof is almost identical to the proof of Lemma 4 by \cite{YinTatPaper}. As in there we will use the following lemma on matrix concentration, which appears as Lemma 11 in the work of \cite{YinTatPaper}.
\begin{lemma}[\cite{YinTatPaper}]
\label{lemma:concbetter}
Let $Y_1 \ldots Y_k$ be independent random positive semidefinite matrices of size $d \times d$. Let $Y = \sum Y_i$ and let $Z = E[Y]$. If $Y_i \preceq R\cdot Z$, then 
\[ P\left[ \sum Y_i \preceq (1 - \epsilon)Z \right] \leq d e^{\frac{\epsilon^2}{2R}}\]
\[ P\left[ \sum Y_i \succeq (1 + \epsilon)Z \right] \leq d e^{\frac{\epsilon^2}{3R}}.\]
\end{lemma}

For every matrix $A_i$ choose $Y_i = \frac{A_i}{p_i}$ with probability $p_i$ and $0$ otherwise. Therefore we need to bound $\sum Y_i$. Note that $E[Y_i] = A$. We will now show that 
\[ \forall i \;\; Y_i \preceq \frac{1}{c\log d \epsilon^{-2}} A\]
which will finish the proof by a direct application of Lemma \ref{lemma:concbetter}. 
First we will show that 
\[ \frac{A_i}{\tau_i(A)} \preceq A.\]
We need to show that $\forall \x \; \x^T A_i \x \leq \tau_i(A) \x^T A \x$. By noting that the $A_i$ are PSD we have that if $\x^TA\x = 0$ then $\forall i,  \x^TA_i\x = 0$. We can now consider $\x = A^{+/2} \y$. Therefore, we need to show that 
\[ \y A^{+/2} A_i A^{+/2} \y \leq \tau_i(A) \|y\|^2.\]
This is true because the maximum eigenvalue of $A^{+/2} A_i A^{+/2}$ is bounded by its \\
$Trace(A^{+/2} A_i A^{+/2})$ (due to positive semidefiniteness), which is equal to $\tau_i(A)$ by definition. 
Therefore when $p_i < 1$, i.e., $\alpha c \bu_i \log(d) < 1$, the facts $\tau_i(A) \leq \bu_i $ immediately provide 
\[ Y_i \preceq \frac{1}{c\log d \epsilon^{-2}} A.\]
When $p_i = 1$ the above does not hold, but we can essentially replace $Y_i = A_i$ with
$c\log d \epsilon^{-2}$ variables each equal to $\frac{A_i}{c\log d \epsilon^{-2}}$, each being sampled with probability 1. This does not change $E[\sum Y_i]$ but proves concentration. Now a direct application of Lemma \ref{lemma:concbetter} finishes the proof. Also note that a standard Chernoff bound proves the required bound on the sample size.
\end{proof}

\subsection{Proof of Lemma \ref{thm:tautildetheorem}} 
\begin{proof}
The proof is almost identical to the proof of Theorem 1 by \cite{YinTatPaper}. We include the proof here for completeness. 
Define $S_i$ to be $S$ if $\exists j$ such that $X_j = A_i$, and $S + A_i$ otherwise. Now since $S^{i} \preceq A$, we have that $\tilde{\tau}_i^S(A) \geq \tau_i(A)$. To bound the expected sum we break it down into two parts:
 \[ \sum_i \tilde{\tau}_i^S(A) = \sum_{i \in S} \tilde{\tau}_i^S(A) + \sum_{i \notin S} \tilde{\tau}_i^S(A).\]
 The first part is a sum of the leverage scores of $S$, and so it is bounded by $d$. To bound the second term, consider a random process that first selects $S$, then selects a random $i \notin S$ and returns $\tilde{\tau}_i^S(A)$. There are always exactly $m-r$ such $i$, so the value returned by this random process is, in expectation, exactly equal to $\frac{1}{m-r} E[\sum_{i \notin S} \tilde{\tau}_i^S(A)]$. 
 This random process is also equivalent to randomly selecting a set $S′$ of size $r + 1$, then
randomly choosing an $i \in S'$ and returning its leverage score. In expectation it is therefore equal to the average leverage score in $S′$. $S'$ has size $r+1$, and its leverage scores sum to its
rank, so we can bound its average leverage score by $\frac{d}{r+1}$. Overall, we have that 
\[ E [\sum_i \tilde{\tau}_i^S(A)] = d + \frac{d(m-r)}{r+1} \leq O\left(\frac{md}{r}\right). \]
\end{proof}

\subsection{Proof of Ellipsoidal Cover Lemma}
\label{sec:ellipsoidcoverappendix}
\begin{proof}[Proof of Lemma \ref{lemma:ConstantDecreaseLemma}] 
Let $\lambda(\x)$ be the Newton decrement at $\x$.
By Lemma \ref{lemma:lambdalemma}, we know that if $\lambda(\x) \leq
\frac{r}{1+r} < 1$, then $$||\x - \x^*||_{\x} \leq
\frac{\lambda(\x)}{1-\lambda(\x)} = r.$$
Consider a single iteration of the algorithm at time $t$. If $\lambda(\x_t) \leq \frac{r}{1+r}$, then we may conclude that  $\x^* \in W_{r}(\x_t)$. Therefore, it is only when $\lambda(\x_t) > \frac{r}{1+r}$ that  $\x^*$ may not be contained within $W_{r}(\x_t)$. Since our update is of the form 
$$\x_{t+1} = \x_t - \frac{1}{\gamma(1+\lambda(\x_t))}\hessinv f(\x_t) \grad f(\x_t),$$
we have the following:
\begin{align*}
f(\x_{t+1}) &\leq f(\x_t) - \frac{1}{\gamma(1+\lambda(\x_t))} \nabla f(\x_t)^{\top}([\nabla^2 f(\x_t)]^{-1}\nabla f(\x_t)) + \rho\bigg(\frac{\lambda(\x_t)}{\gamma(1+\lambda(\x_t))}\bigg)\\
&=  f(\x_t) - \frac{(\lambda(\x_t))^2}{\gamma(1+\lambda(\x_t))} + \rho\bigg(\frac{\lambda(\x_t)}{\gamma(1+\lambda(\x_t))}\bigg)\\
&=  f(\x_t) - \frac{(\lambda(\x_t))^2}{\gamma(1+\lambda(\x_t))} - \ln\bigg(1- \frac{\lambda(\x_t)}{\gamma(1+\lambda(\x_t))}\bigg) - \frac{\lambda(\x_t)}{\gamma(1+\lambda(\x_t))}\\
&= f(\x_t) - \frac{\lambda(\x_t)}{\gamma} + \ln\bigg(1 + \frac{\lambda(\x_t)}{\gamma + (\gamma-1)\lambda(\x_t)}\bigg)
\end{align*}
where the first inequality follows from Lemma \ref{lemma:smoothness}. It now
follows that
\begin{align*}
f(\x_t) - f(\x_{t+1}) &\geq \frac{\lambda(\x_t)}{\gamma} - \ln\bigg(1 + \frac{\lambda(\x_t)}{\gamma + (\gamma-1)\lambda(\x_t)}\bigg)\\
&\geq \frac{\lambda(\x_t)}{\gamma} -  \frac{\lambda(\x_t)}{\gamma + (\gamma-1)\lambda(\x_t)}\\
&= \lambda(\x_t)\bigg(\frac{1}{\gamma} - \frac{1}{\gamma+(\gamma - 1)\lambda(\x_t)}\bigg)\\
&> \frac{r}{1+r}\bigg(\frac{1}{\gamma} - \frac{1}{\gamma+\frac{(\gamma-1)r}{1+r}}\bigg)\\
& > 0
\end{align*} 
where the second inequality comes from the fact that, for all $x \in \reals$, $\ln(1+x) \leq 1+x$. Let $\nu = \frac{r}{1+r}\bigg(\frac{1}{\gamma} - \frac{1}{\gamma+\frac{(\gamma-1)r}{1+r}}\bigg)$.
Then we see that after $$\frac{f(\x_1) - f(\x^*)}{\nu}$$ steps, we can guarantee that we have arrived at $\x_t$ such that $\x^* \in W_{r}(\x_t)$.
\end{proof}

\begin{algorithm}[]
\caption{\textbf{N-SVRG}}
\label{alg:modifiedSVRG}
\begin{algorithmic}
\REQUIRE Convex set $\K$, $f(\x) = \frac{1}{m}\sum\limits_{k = 1}^m f_k(\x)$, Norm $A
\succeq 0$, update frequency $S_1$,  and learning rate $\eta$

\STATE{Initialize $\tilde{\x}_0$}
\FOR{$s = 1,2 \ldots $}
\STATE{$\tilde{\x} = \tilde{\x}_{s-1}$}
\STATE{$\tilde{\mu} = \frac{1}{m}\sum_{k=1}^m \nabla f_k({\tilde{\x}})$}
\STATE{$\x_0 = \tilde{\x}$}
\FOR{$t = 1$ to $S_1$}
\STATE{Randomly choose $i_t \in \{1 \ldots m\}$}
\STATE{$g = \nabla
f_{i_t}(\x_{t-1}) -
\nabla
f_{i_t}(\tilde{\x}) + \tilde{\mu}$}
\STATE{$\x_t = \Pi_{\K}^A \left(\x_{t-1} - \eta A^{-1}g \right)$}
\ENDFOR
\STATE{$\tilde{\x}_s = \x_t$ for randomly chosen $t \in \{1, \ldots ,m\}$}
\ENDFOR

\end{algorithmic}
\end{algorithm}

\subsection{Proof of SVRG Lemma}
\begin{proof}[Proof of Lemma \ref{lemma:SVRGLemma}]
The proof follows the original proof of SVRG \cite{SVRG} with a few
modifications to take the general norm into account.
For any $i$, consider 
\[g_i(\x) = f_i(\x) - f_i(\x^*) - \innerprod{\nabla f_i(\x^*)}{\x - \x^*}.\]
We know that $g_i(\x^*) = \min_{\x} g_i(\x)$ since $\nabla g_i(\x^*) = 0$. Therefore,
\begin{eqnarray*}
0 = g_i(\x^*) & \leq & \min_{\eta}\left( g_i(\x - \eta A^{-1} \nabla g_i(\x^*))
\right)
\\
&\leq& \min_{\eta} \left( g_i(\x) - \eta \invmatnorm{g_i(\x)}{A}^2 + 0.5
\beta \eta^2 \invmatnorm{g_i(\x)}{A}^2 \right)\\
&=& g_i(\x) - \frac{1}{2\beta} \invmatnorm{g_i(\x)}{A}^2\enspace.
\end{eqnarray*} 
The second inequality follows from smoothness of $g_i$, where we note that $g_i$ is as
smooth as $f_i$. Summing the above inequality over $i$ and setting $\x^*$ to be
the minimum of $f$ so that $\nabla f(\x^*) = 0$, we get 
\[ \frac{\sum_{i=1}^{m} \invmatnorm{\nabla f_i(\x) - \nabla
f_i(\x^*)}{A}^2}{2m\beta} \leq f(\x) - f(\x^*).\]
Define $v_t = \nabla f_{i_t}(\x_{t-1}) - \nabla f_{i_t}(\tilde{\x}) +
\tilde{\mu}$. Conditioned on $\x_{t-1}$, we can take the expectation with respect
to $i_t$ and obtain
\begin{eqnarray*}
\av \left[ \invmatnorm{v_t}{A}^2 \right] &\leq& 2 \av \left[ \invmatnorm{\nabla
f_{i_t}(\x_{t-1}) - \nabla f_{i_t}(\x^*)}{A}^2 \right] + 2 \av \left[
\invmatnorm{\nabla f_{i_t}(\tilde{\x}) - \nabla f_{i_t}(\x^*) - \nabla
f(\tilde{\x})}{A}^2 \right]\\
&\leq& 2 \av \left[ \invmatnorm{\nabla f_{i_t}(\x_{t-1}) -
\nabla f_{i_t}(\x^*)}{A}^2 \right] + 2 \av \left[ \invmatnorm{\nabla
f_{i_t}(\tilde{\x}) - \nabla f_{i_t}(\x^*)}{A}^2 \right]\\
&\leq& 4\beta \left( f(\x_{t-1}) - f(\x^*) + f(\tilde{\x}) - f(\x^*) \right).
\end{eqnarray*}
The above inequalities follow by noting the following three facts:
\begin{itemize}
  \item $\invmatnorm{a+b}{A}^2 \leq 2 \invmatnorm{a}{A}^2 + 2
  \invmatnorm{b}{A}^2\enspace.$
  \item $\tilde{\mu} = \nabla f(\tilde{\x})\enspace.$
  \item $\av \left[ \invmatnorm{X - \av X}{A}^2 \right] \leq \av
  \invmatnorm{X}{A}^2\enspace.$
\end{itemize}
Now note that conditioned on $\x_{t-1}$, we have that $\av v_t = \nabla
f(\x_{t-1})$, and so 
\begin{eqnarray*}
\av \matnorm{\x_t - \x^*}{A}^2 &\leq& \matnorm{\x_{t-1} - \eta A^{-1}v_t - \x^*
}{A}^2
\\
&=& \matnorm{\x_{t-1} - \x^*}{A}^2 -
2\eta\innerprod{\x_{t-1} - \x^*}{A A^{-1} \av v_t} + \eta^2 \av
\invmatnorm{v_t}{A}^2
\\
&\leq& \matnorm{\x_{t-1} - \x^*}{A}^2 -
2\eta\innerprod{\x_{t-1} - \x^*}{\nabla f(\x_{t-1})} + 4\beta\eta^2 \left(
f(\x_{t-1}) - f(\x^*) + f(\tilde{\x}) - f(\x^*)\right) \\
&\leq & \matnorm{\x_{t-1} - \x^*}{A}^2 - 2 \eta \left( f(\x_{t-1}) - f(\x^*) \right)
+ 4\beta\eta^2 \left(
f(\x_{t-1}) - f(\x^*) + f(\tilde{\x}) - f(\x^*)\right) \\ 
&=& \matnorm{\x_{t-1} - \x^*}{A}^2 - 2 \eta (1 - 2\eta \beta) \left( f(\x_{t-1}) -
f(\x^*) \right) + 4\beta\eta^2 \left(f(\tilde{\x}) - f(\x^*)\right).
\end{eqnarray*}
The first inequality uses the Pythagorean inequality for norms, the second
inequality uses the derived inequality for $\av \invmatnorm{v_t}{A}^2$, and the
third inequality uses the convexity of $f(\x)$.
Consider a fixed stage $s$, so that $\tilde{\x} = \tilde{\x}_{s-1}$ and
$\tilde{\x}_s$ are selected after all of the updates have been completed. By summing
the previous inequality over $t$ and taking an expectation over the history, we
get 
\begin{eqnarray*}
\av \matnorm{\x_n - \x^*}{A}^2 + 2 \eta (1 - 2\eta \beta)n \av \left[f(\x_{s}) -
f(\x^*) \right] &\leq& \av \matnorm{\x_0 - \x^*}{A}^2 + 4\beta n\eta^2 \av
\left[f(\tilde{\x}) - f(\x^*)\right] \\
&=& \matnorm{\tilde{\x} - \x^*}{A}^2 + 4\beta n\eta^2 \av \left[f(\tilde{\x}) -
f(\x^*)\right] \\
&\leq& 2\frac{\av \left[ f(\tilde{\x}) - f(\x^*) \right]}{\alpha} + 4\beta n\eta^2
\av \left[f(\tilde{\x}) - f(\x^*)\right] \\
&=& 2\left( \frac{1}{\alpha} + 2\beta n \eta^2\right)\av \left[f(\tilde{\x}) -
f(\x^*)\right].
\end{eqnarray*}
The second inequality uses the strong convexity property. Therefore, we have that
\[ \av \left[f(\x_{s}) -
f(\x^*) \right]  \leq \left( \frac{1}{\alpha\eta (1 - 2\eta \beta)n} +
\frac{2\eta \beta}{(1 - 2\eta \beta)}\right) \av \left[f(\x_{s-1}) -
f(\x^*)\right].  \]
\end{proof}

\ENDDOC

%% file: defs.tex
\DeclareMathOperator*{\argmin}{argmin}

\newcommand{\bb}[1]{$\ll$\textsf{\color{green} Brian : #1}$\gg$}

\def\reals{{\mathbb R}}

\def\norm#1{\mathopen\| #1 \mathclose\|}

\newcommand{\ignore}[1]{}

\def\reals{{\mathbb R}}

\def\bold0{\mathbf{0}}

\def\bb{\mathbf{b}}

\def\bh{\mathbf{h}}

\def\bu{\mathbf{u}}

\def\bu{\mathbf{u}}
\def\bv{\mathbf{v}}

\def\bw{\mathbf{w}}



\newcommand{\K}{\ensuremath{\mathcal K}}

\newcommand{\x}{\ensuremath{\mathbf x}}

\newcommand{\y}{\ensuremath{\mathbf y}}
\newcommand{\z}{\ensuremath{\mathbf z}}
\newcommand{\h}{\ensuremath{\mathbf h}}

\newcommand{\vv}{\ensuremath{\mathbf v}}

\newcommand\pr{\mbox{\bf Pr}}
\newcommand\av{\mbox{\bf E}}

\def\bb{\mathbf{b}}

\def\bu{\mathbf{u}}

\def\bv{\mathbf{v}}

\def\epsilon{\varepsilon}

\def\hestinv{\tilde{\nabla}^{-2}}
\def\hest{\tilde{\nabla}^{2}}
\def\hessinv{\nabla^{-2}}
\def\hess{\nabla^2}
\def\grad{\nabla}
\newcommand{\defeq}{\triangleq}

\newcommand{\innerprod}[2]{\langle #1, #2 \rangle}
\newcommand{\matnorm}[2]{\| #1 \|_{#2}}
\newcommand{\invmatnorm}[2]{\| #1 \|_{#2^{-1}}}

\newtheorem{theorem}{Theorem}[section]
\newtheorem*{theorem*}{Theorem}

\newtheorem{lemma}[theorem]{Lemma}
\newtheorem{corollary}[theorem]{Corollary}

\newtheorem{definition}[theorem]{Definition}

\newtheorem{remark}[theorem]{Remark}

\newtheorem{fact}[theorem]{Fact}

%% file: main.bbl
\newcommand{\etalchar}[1]{$^{#1}$}
\begin{thebibliography}{XYRK{\etalchar{+}}16}

\bibitem[AAZB{\etalchar{+}}17]{agarwal2016finding}
Naman Agarwal, Zeyuan Allen-Zhu, Brian Bullins, Elad Hazan, and Tengyu Ma.
\newblock Finding approximate local minima faster than gradient descent.
\newblock In {\em Proceedings of the 49th Annual ACM SIGACT Symposium on Theory
  of Computing}, pages 1195--1199. ACM, 2017.

\bibitem[AS16]{arjevani2016oracle}
Yossi Arjevani and Ohad Shamir.
\newblock Oracle complexity of second-order methods for finite-sum problems.
\newblock {\em arXiv preprint arXiv:1611.04982}, 2016.

\bibitem[AZ16]{Katyusha2016}
Zeyuan Allen-Zhu.
\newblock Katyusha: The first direct acceleration of stochastic gradient
  methods.
\newblock {\em arXiv preprint arXiv:1603.05953}, 2016.

\bibitem[BBN16]{bollapragada2016exact}
Raghu Bollapragada, Richard Byrd, and Jorge Nocedal.
\newblock Exact and inexact subsampled {Newton} methods for optimization.
\newblock {\em arXiv preprint arXiv:1609.08502}, 2016.

\bibitem[BCNN11]{BCNN11}
Richard~H. Byrd, Gillian~M. Chin, Will Neveitt, and Jorge Nocedal.
\newblock On the use of stochastic {Hessian} information in optimization
  methods for machine learning.
\newblock {\em SIAM Journal on Optimization}, 21(3):977--995, 2011.

\bibitem[BD99]{covertype}
Jock~A. Blackard and Denis~J. Dean.
\newblock Comparative accuracies of artificial neural networks and discriminant
  analysis in predicting forest cover types from cartographic variables.
\newblock {\em Computers and Electronics in Agriculture}, 24(3):131--151, 1999.

\bibitem[BHNS16]{BHNS14}
Richard~H. Byrd, Samantha~L. Hansen, Jorge Nocedal, and Yoram Singer.
\newblock A stochastic quasi-{Newton} method for large-scale optimization.
\newblock {\em SIAM Journal on Optimization}, 26(2):1008--1031, 2016.

\bibitem[Bro70]{bro70}
Charles~G. Broyden.
\newblock The convergence of a class of double-rank minimization algorithms: 2.
  the new algorithm.
\newblock {\em IMA Journal of Applied Mathematics}, 6(3):222--231, 1970.

\bibitem[CLM{\etalchar{+}}15]{YinTatPaper}
Michael~B. Cohen, Yin~Tat Lee, Cameron Musco, Christopher Musco, Richard Peng,
  and Aaron Sidford.
\newblock Uniform sampling for matrix approximation.
\newblock In {\em Proceedings of the 2015 Conference on Innovations in
  Theoretical Computer Science}, pages 181--190. ACM, 2015.

\bibitem[Coh16]{cohensubspaceembeddings}
Michael~B. Cohen.
\newblock Nearly tight oblivious subspace embeddings by trace inequalities.
\newblock In {\em Proceedings of the Twenty-Seventh Annual ACM-SIAM Symposium
  on Discrete Algorithms}, pages 278--287. Society for Industrial and Applied
  Mathematics, 2016.

\bibitem[DBLJ14]{SAGA}
Aaron Defazio, Francis Bach, and Simon Lacoste-Julien.
\newblock Saga: A fast incremental gradient method with support for
  non-strongly convex composite objectives.
\newblock In {\em Advances in Neural Information Processing Systems}, pages
  1646--1654, 2014.

\bibitem[DHS11]{adagrad}
John Duchi, Elad Hazan, and Yoram Singer.
\newblock Adaptive subgradient methods for online learning and stochastic
  optimization.
\newblock {\em Journal of Machine Learning Research}, 12(Jul):2121--2159, 2011.

\bibitem[EM15]{newsamp}
Murat~A. Erdogdu and Andrea Montanari.
\newblock Convergence rates of sub-sampled {Newton} methods.
\newblock In {\em Advances in Neural Information Processing Systems}, pages
  3034--3042, 2015.

\bibitem[Fle70]{fle70}
Roger Fletcher.
\newblock A new approach to variable metric algorithms.
\newblock {\em The Computer Journal}, 13(3):317--322, 1970.

\bibitem[Gol70]{gol70}
Donald Goldfarb.
\newblock A family of variable-metric methods derived by variational means.
\newblock {\em Mathematics of Computation}, 24(109):23--26, 1970.

\bibitem[JZ13]{SVRG}
Rie Johnson and Tong Zhang.
\newblock Accelerating stochastic gradient descent using predictive variance
  reduction.
\newblock In {\em Advances in Neural Information Processing Systems}, pages
  315--323, 2013.

\bibitem[LACBL16]{luo2016efficient}
Haipeng Luo, Alekh Agarwal, Nicol{\`o} Cesa-Bianchi, and John Langford.
\newblock Efficient second order online learning by sketching.
\newblock In {\em Advances in Neural Information Processing Systems}, pages
  902--910, 2016.

\bibitem[LC98]{mnist}
Yann LeCun and Corinna Cortes.
\newblock The {MNIST} database of handwritten digits.
\newblock 1998.

\bibitem[Lic13]{uci}
Moshe Lichman.
\newblock {UCI} machine learning repository, 2013.

\bibitem[LLX14]{lin2014accelerated}
Qihang Lin, Zhaosong Lu, and Lin Xiao.
\newblock An accelerated proximal coordinate gradient method.
\newblock In {\em Advances in Neural Information Processing Systems}, pages
  3059--3067, 2014.

\bibitem[LMH15]{Catalyst2015}
Hongzhou Lin, Julien Mairal, and Za{\"{\i}}d Harchaoui.
\newblock A universal catalyst for first-order optimization.
\newblock In {\em Advances in Neural Information Processing Systems}, pages
  3384--3392, 2015.

\bibitem[LMP13]{PengRowSampling}
Mu~Li, Gary~L. Miller, and Richard Peng.
\newblock Iterative row sampling.
\newblock In {\em IEEE 54th Annual Symposium on Foundations of Computer
  Science}, pages 127--136. IEEE, 2013.

\bibitem[Mar10]{MAR10}
James Martens.
\newblock Deep learning via {Hessian}-free optimization.
\newblock In {\em International Conference on Machine Learning}, pages
  735--742, 2010.

\bibitem[McC97]{realsim}
Andrew McCallum.
\newblock Real-sim, 1997.
\newblock Available at
  \url{https://www.csie.ntu.edu.tw/~cjlin/libsvmtools/datasets/binary.html#real-sim}.

\bibitem[MNJ16]{SLBFGS}
Philipp Moritz, Robert Nishihara, and Michael Jordan.
\newblock A linearly-convergent stochastic {L-BFGS} algorithm.
\newblock In {\em Artificial Intelligence and Statistics}, pages 249--258,
  2016.

\bibitem[MR14]{MRSQN}
Aryan Mokhtari and Alejandro Ribeiro.
\newblock {RES:} regularized stochastic {BFGS} algorithm.
\newblock {\em IEEE Transactions on Signal Processing}, 62(23):6089--6104,
  2014.

\bibitem[Nem04]{NemirovskiBook}
Arkadi Nemirovski.
\newblock Interior point polynomial time methods in convex programming.
\newblock {\em Lecture notes}, 2004.

\bibitem[Nes83]{nesterovacceleration}
Yurii Nesterov.
\newblock A method of solving a convex programming problem with convergence
  rate {O}(1/k2).
\newblock In {\em Soviet Mathematics Doklady}, volume~27, pages 372--376, 1983.

\bibitem[Nes13]{NesterovBook}
Yurii Nesterov.
\newblock {\em Introductory Lectures on Convex Optimization: A Basic Course}.
\newblock Springer Science \& Business Media, 2013.

\bibitem[NW06]{nocedalbook}
Jorge Nocedal and Stephen Wright.
\newblock {\em Numerical Optimization}.
\newblock Springer Science \& Business Media, 2006.

\bibitem[PW15]{newtonsketch}
Mert Pilanci and Martin~J. Wainwright.
\newblock {Newton} sketch: A linear-time optimization algorithm with
  linear-quadratic convergence.
\newblock {\em arXiv preprint arXiv:1505.02250}, 2015.

\bibitem[RM51]{RM51}
Herbert Robbins and Sutton Monro.
\newblock A stochastic approximation method.
\newblock {\em The Annals of Mathematical Statistics}, pages 400--407, 1951.

\bibitem[RSB12]{Bachpaper}
Nicolas~L. Roux, Mark Schmidt, and Francis~R. Bach.
\newblock A stochastic gradient method with an exponential convergence rate for
  finite training sets.
\newblock In {\em Advances in Neural Information Processing Systems}, pages
  2663--2671, 2012.

\bibitem[Sha70]{sha70}
David~F. Shanno.
\newblock Conditioning of quasi-{Newton} methods for function minimization.
\newblock {\em Mathematics of Computation}, 24(111):647--656, 1970.

\bibitem[SSZ13]{SDCA}
Shai Shalev-Shwartz and Tong Zhang.
\newblock Stochastic dual coordinate ascent methods for regularized loss
  minimization.
\newblock {\em Journal of Machine Learning Research}, 14(Feb):567--599, 2013.

\bibitem[SSZ14]{dane}
Ohad Shamir, Nathan Srebro, and Tong Zhang.
\newblock Communication-efficient distributed optimization using an approximate
  {Newton}-type method.
\newblock In {\em International Conference on Machine Learning}, pages
  1000--1008, 2014.

\bibitem[SSZ16]{AccSDCA}
Shai Shalev-Shwartz and Tong Zhang.
\newblock Accelerated proximal stochastic dual coordinate ascent for
  regularized loss minimization.
\newblock {\em Mathematical Programming}, 155(1-2):105--145, 2016.

\bibitem[SYG07]{SYGSQN}
Nicol~N. Schraudolph, Jin Yu, and Simon G{\"u}nter.
\newblock A stochastic quasi-{Newton} method for online convex optimization.
\newblock In {\em Artificial Intelligence and Statistics}, pages 436--443,
  2007.

\bibitem[Tro12]{Tropp}
Joel~A Tropp.
\newblock User-friendly tail bounds for sums of random matrices.
\newblock {\em Foundations of Computational Mathematics}, 12(4):389--434, 2012.

\bibitem[XYRK{\etalchar{+}}16]{xu2016sub}
Peng Xu, Jiyan Yang, Farbod Roosta-Khorasani, Christopher R{\'e}, and
  Michael~W. Mahoney.
\newblock Sub-sampled {Newton} methods with non-uniform sampling.
\newblock In {\em Advances in Neural Information Processing Systems}, pages
  3000--3008, 2016.

\bibitem[YLZ17]{ye2017unifying}
Haishan Ye, Luo Luo, and Zhihua Zhang.
\newblock A unifying framework for convergence analysis of approximate {Newton}
  methods.
\newblock {\em arXiv preprint arXiv:1702.08124}, 2017.

\bibitem[ZMJ13]{mahdavi}
Lijun Zhang, Mehrdad Mahdavi, and Rong Jin.
\newblock Linear convergence with condition number independent access of full
  gradients.
\newblock In {\em Advances in Neural Information Processing Systems}, pages
  980--988, 2013.

\end{thebibliography}
